\documentclass[12pt]{article}

\usepackage{color}
\usepackage{graphicx}
\usepackage{epstopdf}
\usepackage{caption}
\usepackage{multicol}
\usepackage{lineno}
\usepackage{ulem}
\usepackage{url}
\usepackage{multirow}
\usepackage{algorithm, algorithmic}
\usepackage{longtable}
\usepackage[linkcolor=blue]{hyperref}
\usepackage{threeparttable}
\usepackage{verbatim}
\usepackage{lineno}
\usepackage{amssymb,mathrsfs,amsmath}

\makeatletter
\def\UrlAlphabet{%
      \do\a\do\b\do\c\do\d\do\e\do\f\do\g\do\h\do\i\do\j%
      \do\k\do\l\do\m\do\n\do\o\do\p\do\q\do\r\do\s\do\t%
      \do\u\do\v\do\w\do\x\do\y\do\z\do\A\do\B\do\C\do\D%
      \do\E\do\F\do\G\do\H\do\I\do\J\do\K\do\L\do\M\do\N%
      \do\O\do\P\do\Q\do\R\do\S\do\T\do\U\do\V\do\W\do\X%
      \do\Y\do\Z}
\def\UrlDigits{\do\1\do\2\do\3\do\4\do\5\do\6\do\7\do\8\do\9\do\0}
\g@addto@macro{\UrlBreaks}{\UrlOrds}
\g@addto@macro{\UrlBreaks}{\UrlAlphabet}
\g@addto@macro{\UrlBreaks}{\UrlDigits}
\makeatother

\usepackage{scicite}

\usepackage{times}

\newenvironment{sciabstract}{%
\begin{quote} \bf}
{\end{quote}}

\title{OpenClinicalAI: enabling AI to diagnose diseases in real-world clinical settings}

\author{Yunyou Huang,$^{1,6}$ Nana Wang,$^{2,5}$ Suqin Tang,$^{1,6}$ Li Ma,$^{3,6}$ Tianshu Hao,$^{2,5}$ \\ Zihan Jiang,$^{2,5}$ Fan Zhang,$^{2,5}$ Guoxin Kang,$^{2,5}$ Xiuxia Miao,$^{1,6}$ Xianglong Guan,$^{1,6}$ \\ Ruchang Zhang,$^{1,6}$ Zhifei Zhang,$^{4,\star}$ Jianfeng Zhan,$^{2,5,6,\star}$ \\ for the Alzheimer's Disease Neuroimaging Initiative\footnote{Data used in preparation of this article were obtained from the Alzheimer's Disease Neuroimaging Initiative (ADNI) database (\url{http://adni.loni.usc.edu}). As such, the investigators within the ADNI contributed to the design and implementation of ADNI and/or provided data but did not participate in analysis or writing of this report. A complete listing of ADNI investigators can be found at: \url{http://adni.loni.usc.edu/wp-content/uploads/how_to_apply/ADNI_Acknowledgement_List.pdf}}\\
\\
\normalsize{$^{1}$Guangxi Key Lab of Multi-Source Information Mining \& Security, }\\
\normalsize{School of Computer Science and Engineering \& School of Software, }\\
\normalsize{Guangxi Normal University, Guilin, China}\\
\normalsize{$^{2}$State Key Laboratory of Computer Architecture, Institute of Computing Technology, }\\
\normalsize{Chinese Academy of Sciences, Beijing, China}\\
\normalsize{$^{3}$Guilin Medical University, Guilin, China}\\
\normalsize{$^{4}$Department of Physiology and Pathophysiology, Capital Medical University, Beijing, China}\\
\normalsize{$^{5}$University of Chinese Academy of Sciences, China}\\
\normalsize{$^{6}$International Open Benchmark Council}\\
\\
\normalsize{$^\star$To whom correspondence should be addressed;}\\
\normalsize{E-mail:  zhanjianfeng@ict.ac.cn or zhifeiz@ccmu.edu.cn}
}

\date{}

\begin{document} 


\baselineskip24pt


\maketitle

%. Those conditions are too ideal to implementing those AI systems 
% where there are unknown and unfamiliar diseases

\begin{sciabstract}
%(The abstract should be 125 words or less.)

This paper quantitatively reveals the state-of-the-art and state-of-the-practice AI systems only achieve acceptable performance on the stringent conditions that all categories of subjects are known, which we call closed clinical settings, but fail to work
in real-world clinical settings. Compared to the diagnosis task in the closed setting, real-world clinical settings pose severe challenges, and we must treat them differently. We build a clinical AI benchmark named Clinical AIBench to set up real-world clinical settings to facilitate %real-world clinical AI 
researches. We propose an open, dynamic machine learning framework and develop an AI system named OpenClinicalAI to diagnose diseases in real-world clinical settings. The first versions of Clinical AIBench and OpenClinicalAI target Alzheimer's disease. In the real-world clinical setting, OpenClinicalAI significantly outperforms the state-of-the-art AI system. In addition, OpenClinicalAI develops personalized diagnosis strategies to avoid unnecessary testing and seamlessly collaborates with clinicians. It is promising to be embedded in the current medical systems to improve medical services.
%Real-world clinical settings are open with uncertainty and complexity,  It infinitely expands the scale of diagnosis task, and changes the essence of diagnosis task forming a new task for AI. We propose an open, dynamic machine learning framework and develop an AI system to diagnose Alzheimer’s disease in real-world clinical settings. Experiments demonstrate that solving well the diagnosis task on stringent conditions is not much help to solve the diagnosis task in the real-world clinical setting and our AI system significantly outperforms the state-of-the-art AI system. Further, our AI system develops personalized diagnosis strategies to avoid unnecessary testing and seamlessly collaborates with the clinicians. 
\end{sciabstract}

\textbf{One-Sentence Summary:} 
%Real-world clinical settings pose severe challenges to state-of-the-art and state-of-the-practice AI systems. 
We propose a clinical AI benchmark and an open, dynamic machine learning framework to enable AI diagnosis systems to land in real-world clinical settings.
%(These should be a maximum of 125 characters and should complement rather than repeat the title)
%to diagnose diseases in real-world clinical settings. 
% In setting up this template for *Science* papers, we've used both
% the \section* command and the \paragraph* command for topical
% divisions.  Which you use will of course depend on the type of paper
% you're writing.  Review Articles tend to have displayed headings, for
% which \section* is more appropriate; Research Articles, when they have
% formal topical divisions at all, tend to signal them with bold text
% that runs into the paragraph, for which \paragraph* is the right
% choice.  Either way, use the asterisk (*) modifier, as shown, to

\section*{Introduction}

\begin{figure*}
\centering
\begin{minipage}[a]{0.48\textwidth}
%\caption*{a.}
\leftline{\textbf{a}}
\includegraphics[height=7.5cm, width=5cm]{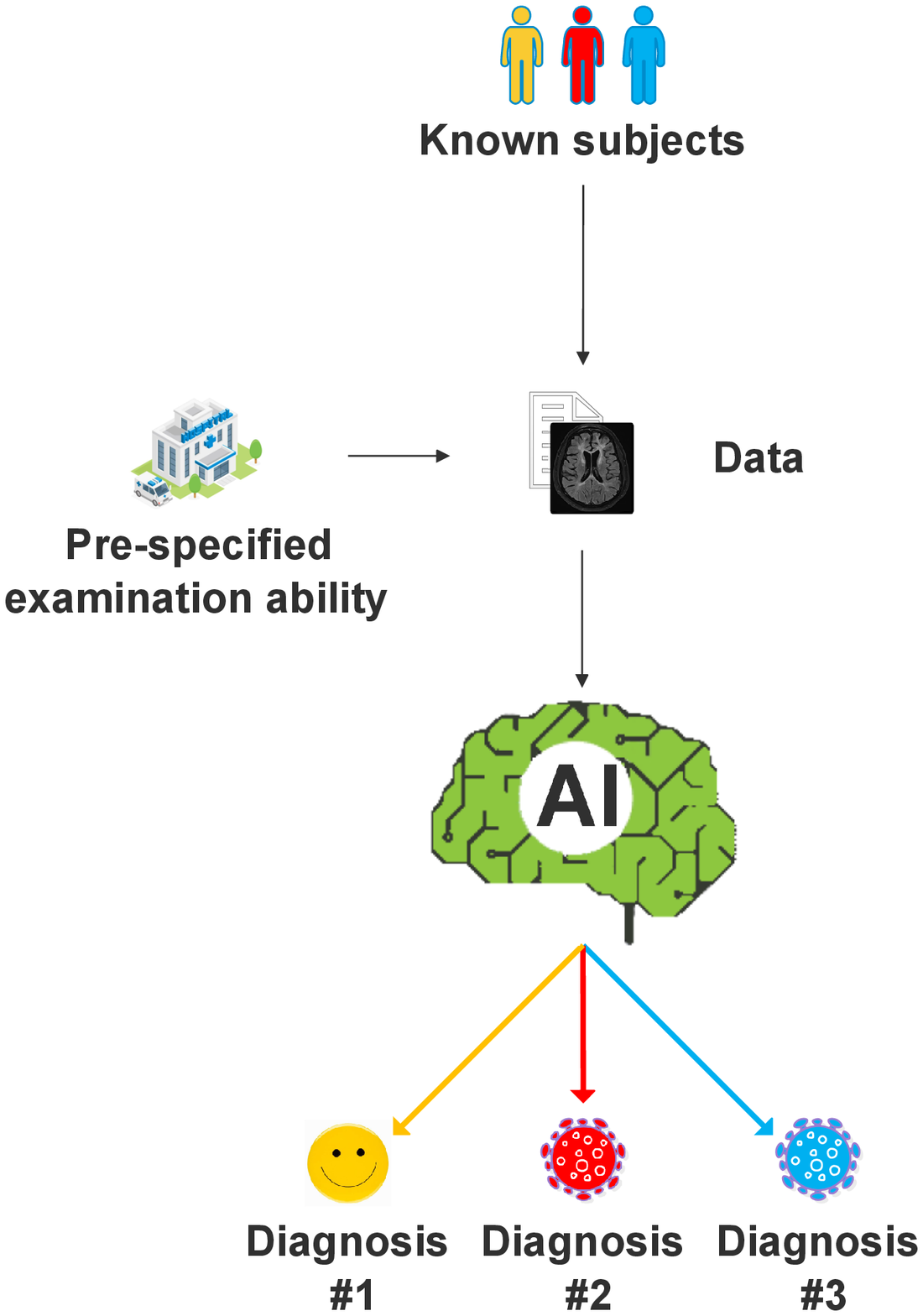} 
\end{minipage}
%\hspace{0.5cm}
\begin{minipage}[a]{0.48\textwidth}
\leftline{\textbf{b}}
\includegraphics[height=7.5cm, width=8.3cm]{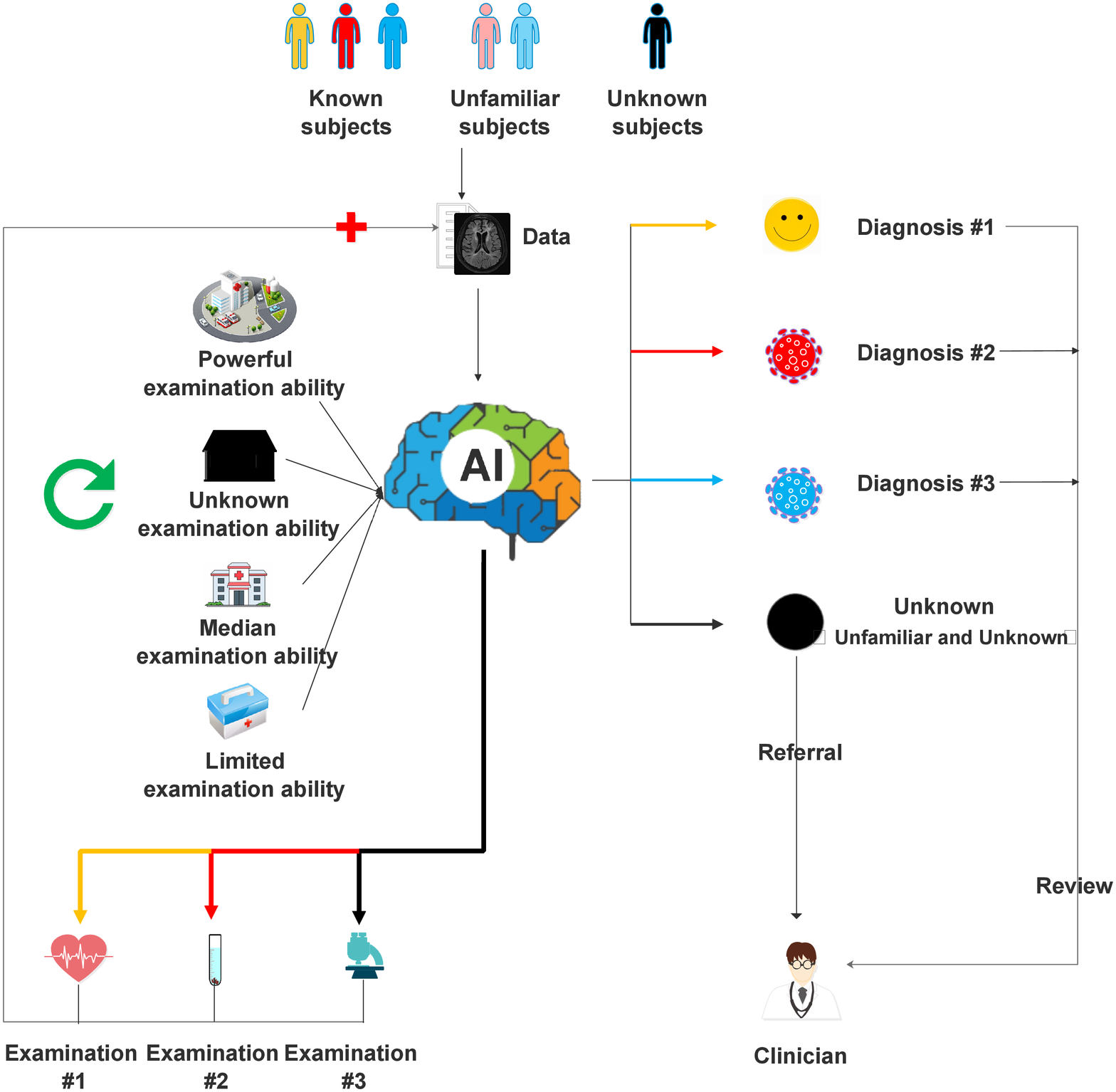} 
\end{minipage}\\
\quad
\caption{\textbf{The workflow of the baseline clinical AI system and OpenClinicalAI. a,} The workflow of the mainstream AI-based diagnostic systems for closed settings. The system only accepts subjects with pre-specified clinical states. First, the same pre-specified medical examinations will be executed by a medical institution with the pre-specified examination ability for every subject. And then, the system will calculate the probability of each pre-defined clinical state for the subject according to the examination. Finally, the system will take the clinical states with the maximum probability as the output and make the final diagnosis. \textbf{b,} The workflow of OpenClinicalAI. It can deal with different categories of subjects, including the unfamiliar and unknown categories of subjects during the development of the system. It can deploy in various medical institutions with different examination abilities from small-scale country clinics to large-scale hospitals. First, OpenClinicalAI will obtain the basic information of the subject and combine the history clinical information of the subject as input. Second, according to the input, OpenClinicalAI calculates the probability of each disease-related examination and each pre-defined clinical state, including the unknown clinical state. Third, for each pre-defined clinical state, if a clinical state's possibility is greater than the specific threshold, then the clinical state is the final diagnosis of OpenClinicalAI, which will be sent to clinicians to review. Otherwise, go to the next step. Fourth, for each examination, if the probability of an examination is greater than the specified threshold and the medical institution can execute this examination, then obtain the examination data, add the data to the input of OpenClinicalAI, and go to step two. Otherwise, go to the next step. Fifth, for the medical institution with specific examination ability, select an executable routine examination with the least cost that has not been executed for the subject, add the examination data to the input of OpenClinicalAI, and go to step two. Otherwise,  go to the next step. Finally, mark the subject without diagnosis as unknown and send them to clinicians to diagnose. Notably, the atypical subject with specified clinical states, unfamiliar and unknown subjects are marked as unknown and sent to the clinician for diagnosis.
\label{fig1}}
\end{figure*}

Due to previous successive successes of AI in the clinical research field, AI is considered a promising technology to provide high-quality and low-cost diagnostic services~\cite{esteva2017dermatologist,mckinney2020international,kermany2018identifying,de2018clinically,ning2018classifying,tang2019interpretable,lian2020attention}. However, there is little evidence that these researches can be implemented into real-world clinical settings (in short, real-world settings) and improve medical services~\cite{he2019practical,brocklehurst2017computerised,roberts2021common}.
 Fig.~\ref{fig1},~\ref{performance_closed},~\ref{performance_open} qualitatively and quantitatively reveal the state-of-the-art and state-of-the-practice AI systems only achieve acceptable performance on the stringent conditions.  We call those stringent conditions closed clinical settings (in short, closed settings).
  The closed  settings have the following primary assumptions: all categories of subjects are known a priori~\cite{bendale2015towards}; the same diagnostic strategy is applied to all subjects, e.g., every subject requires a nuclear magnetic resonance scan (MRI)~\cite{qiu2020development}; the state-of-the-art AI systems can only be deployed at medical institutions that are able to execute the pre-prescribed diagnostic strategy~\cite{de2018clinically,titano2018automated,lee2019explainable}. Vice versa, if the medical institution can not meet prerequisite conditions that are able to complete the pre-prescribed diagnostic strategy, the corresponding AI system can not be deployed. In this context, the diagnosis problem is a closed set recognition problem that is artificially simplified~\cite{kermany2018identifying,de2018clinically,lee2019explainable,ning2018classifying,mei2020artificial}.

Close settings are too ideal for real-world  settings.
The real-world  setting is open with uncertainty and complexity.  
The subject in real-world  settings is not all pre-known categories but contains many unknown and unfamiliar categories. 
 Every subject is different, and there is no one-size-fits-all diagnosis strategy. 
Conditions of medical institutions are different and not pre-known, e.g., some hospitals have positron emission tomography (PET). In contrast, most of the other hospitals in underdeveloped areas are not equipped with PET. The diagnosis problem in real-world settings is an open set recognition problem~\cite{geng2020recent}.

Essentially, the diagnosis task in the closed setting is to find the optimal solution to classify different categories of subjects in a limited space (so-called supervised task) with the help of the ground truth of every subject. However, the real-world setting is open and puts the diagnosis task into unlimited space. Compared to the limited space of closed settings, the infinite space of real-world settings infinitely expands the scale of solving. Moreover, supervised learning will lose efficacy since some categories of subjects and their ground truth are unknown during the development of the AI model. Hence, the main problem of the diagnosis task in the real-world setting converts to efficiently locate the known subjects from the uncertain and complex real-world setting. Moreover, as shown in Fig.~\ref{performance_closed}a and \ref{performance_open}b,c, solving well the diagnosis task in the closed setting is not much help to solve the diagnosis task in the real-world setting. Compared to the diagnosis task in the closed setting, the diagnosis task in the real-world setting is a new and challenging task that we must treat differently. 

This paper calls for turning medical AI attention from algorithmic research in closed settings to systematic study in real-world settings. Specifically, we construct a clinical AI benchmark named Clinical AIBench, which contains real-world and closed settings to promote the landing of AI in real-world settings. To tackle uncertainty and complexity in real-world settings, we propose an open, dynamic machine learning framework ( Fig. S1) and a diagnostic system named OpenClinicalAI to embed in the current healthcare systems as shown in Fig.~\ref{fig1}b. 

The first versions of Clinical AIBench and OpenClinicalAI target Alzheimer's disease (AD) as AD is an incurable disease that brings a heavy burden to our society (the total payment for individuals with AD or other dementias is estimated at $277$ billion)~\cite{hebert2001annual,hebert2013alzheimer,alzheimer20182018,frigerio2019major}. Early and accurate AD diagnosis will result in the correct management of AD or other dementias, saving up to $\$7.9$ trillion in medical and care costs~\cite{alzheimer20182018}. However, it is estimated that $28$ million of the world's $36$ million people with dementia do not receive a diagnosis since the limited medical resources and experts, etc.~\cite{prince2018world}.

The current version of Clinical AIBench includes two clinical settings, which are curated from a large enriched dataset Alzheimer's disease neuroimaging initiative (ADNI): a closed  setting and a real-world  setting~\cite{mueller2005alzheimer}. OpenClinicalAI is composed of multiple independent parts, which can cooperate to handle unknown subjects in real-world  settings, and dynamically adjust diagnosis strategies according to specific subjects and medical institutions. OpenClinicalAI provides an opportunity to embed the AI-based diagnostic system into the current healthcare systems to cooperate with clinicians to improve healthcare services.

In the real-world  setting of Clinical AIBench, we evaluate the performance of OpenClinicalAI 
against the state-of-the-art AI diagnosis system.
Our evaluations show that the performance of OpenClinicalAI exceeds that of the state-of-the-art AI diagnosis system in the real-world  setting. Additionally, OpenClinicalAI can develop personalized diagnosis strategies for every subject in the real-world  setting, maximizing the patient benefit. 

\section*{Results}
\subsection*{Clinical AIBench}

Clinical AIBench contains real-world and closed  settings to develop and evaluate the AI system designed for real-world  settings. The first version targets Alzheimer's disease. In this section, we focus on real-world  settings.

The diagnosis in a real-world  setting requires clinicians to use both individual clinical expertise and the best available external evidence, which is usually obtained by clinical examination, to make a clinical decision for every specific subject~\cite{sackett1996evidence}. It means that at least two main factors must be considered in the diagnosis task in real-world  settings: the subject and the available clinical examination in the medical institution.

As shown in Fig. S2, the real-world  setting is open with uncertainty and complexity.  The primary characteristics of the real-world  setting are as follows:
\begin{itemize}
\item[(1)] Real-world  settings are open,  and clinicians or AI systems often refer to unknown and unfamiliar categories. Thus, the subject's categories are not all pre-known and familiar. A clinician has different expertise and may be unfamiliar with some diseases. In the real-world setting of Clinical AIBench,  an unknown subject category means that it is not familiar to the clinician or AI system. Thus, we mark both unknown categories and unfamiliar categories as unknown. In this work, Clinical AIBench divides all mild cognitive impairment (MCI) and significant memory concern (SMC) subjects into the test set, which are unknown categories during the development of the AI system. 
%collects subjects with varying conditions 
\item[(2)] Subjects in real-world  settings are under different  situations. In this work, subjects with varying conditions are from 67 sites in two countries ( Table S1). For every subject, data of all visits are included in Clinical AIBench ( Table S2). The interval between two contiguous visits of a subject is usually more than six months.
\item[(3)] Medical institutions in real-world  settings have wildly different executive abilities of the examination. Not all the specific medical institutions and their specific executive abilities of the examination are pre-known. In this work, missing data for subjects are not be filled in the real-world  setting of Clinical AIBench. In the real world, most of the subjects do not have all examination data categories. The purpose of the lack of specific category examination data is to keep the varied executive ability of the examination in different medical institutions. That is to say, in the real-world  setting of Clinical AIBench, the lack of specific category examination data indicates that a medical institute lacks that examination ability.
\end{itemize}

Specifically, in this work, the examination data in ADNI is divided into 13 categories: base information (Base), cognition information (Cog), cognition testing (CE), neuropsychiatric information (Neur), function and behavior information (FB), physical neurological examination (PE), blood testing (Blood), urine testing (Urine), nuclear magnetic resonance scan (MRI), positron emission computed tomography scan with 18-FDG (FDG), positron emission computed tomography scan with AV45 (AV45), gene analysis (Gene), and cerebral spinal fluid analysis (CSF).

Details of the dataset in the real-world setting are as follows. 
\begin{itemize}
\item[(1)]All subjects with labels in ADNI are included. % in this work. 
\item[(2)]85\% AD and cognitively normal (CN) subjects are divided as the training set. 5\% of AD and CN subjects are divided as the validation set. 20\% AD and CN subjects, 100\% MCI subjects, and 100\% SMC are divided as the test set. 
\item[(3)]For every subject, different diagnosis strategies are combined according to the presence of different examination data, and the data of each diagnosis strategy forms a sample. 
\end{itemize}
The test set is not accessible during the training of the AI system. In addition, since each subject may have multiple visits ( each visit of the subject is treated as an independent subject), we stipulate that each subject's visit data can only appear in one of the training set, validation set, and test set.

Sine previous AD diagnosis researches are developed in closed  settings, the closed  setting in Clinical AIBench is similar to the previous research~\cite{li2019deep,choi2019deep,zhou2019latent,ebrahimighahnavieh2020deep,tanveer2020machine,qiu2020development,sharma2021faf,tanveer2021classification,ding2019deep}. Only AD and CN subjects are included in the closed  setting, and only the nuclear magnetic resonance instrument and historical medical records are available. 80\% of subjects are divided as the training set, 5\% of subjects are divided as the validation set, and 15\% of subjects are divided as the test set.

\subsection*{The performance of OpenClinicalAI on Alzheimer's disease diagnosis}

Ebrahimighahnavieh et al. and Tanveer et al. review many important works of Alzheimer's disease diagnosis~\cite{ebrahimighahnavieh2020deep,tanveer2020machine}. Most of these works are based on MRI data and transfer learning obtain the most excellent results. In addition, among the recent AI diagnosis researches, the transfer learning framework of the pre-trained model followed by a classifier achieves the state-of-the-art performance in many diagnosis tasks based on medical images~\cite{lee2019explainable,esteva2017dermatologist,tschandl2020human,poplin2018prediction,kermany2018identifying}. Thus, based on the state-of-the-art transfer learning framework and MRI data, we utilize a trained model named DenseNet201~\cite{huang2017densely} and a classifier called XGBoot~\cite{chen2016xgboost} to develop an Alzheimer's disease diagnosis AI system, which we consider as the baseline system to compare against OpenClinicalAI in the rest of this paper. 

We validate the effectiveness of OpenClinicalAI in two ways. First, we compare OpenClinicalAI to the baseline system 
%that uses transfer learning to deal with the MRI and historical data of the subject 
in the closed  setting. Second, we compare OpenClinicalAI to the baseline  system in the real-world setting. Our comparison metrics are the area under the receiver operating characteristic (ROC) curve (AUC) and sensitivity. The larger the value of AUC and sensitivity are, the better the AI system is.

\subsection*{The performance of OpenClinicalAI against the baseline system in the closed setting.}
To the best of our knowledge, all state-of-the-art and state-of-the-practice Alzheimer's disease diagnosis AI researches are developed and evaluated in closed  settings~\cite{ebrahimighahnavieh2020deep,tanveer2020machine,qiu2020development,sharma2021faf,tanveer2021classification,ding2019deep}. We firstly assess the baseline AI system in the closed  setting, and then evaluate OpenClinicalAI in the same closed  setting without the limitation of that only the nuclear magnetic resonance instrument and historical medical records are available.

\begin{figure*}
\centering
\begin{minipage}[a]{0.48\textwidth}
%\caption*{a.}
\leftline{\textbf{a}}
\includegraphics[height=5.5cm, width=7.5cm]{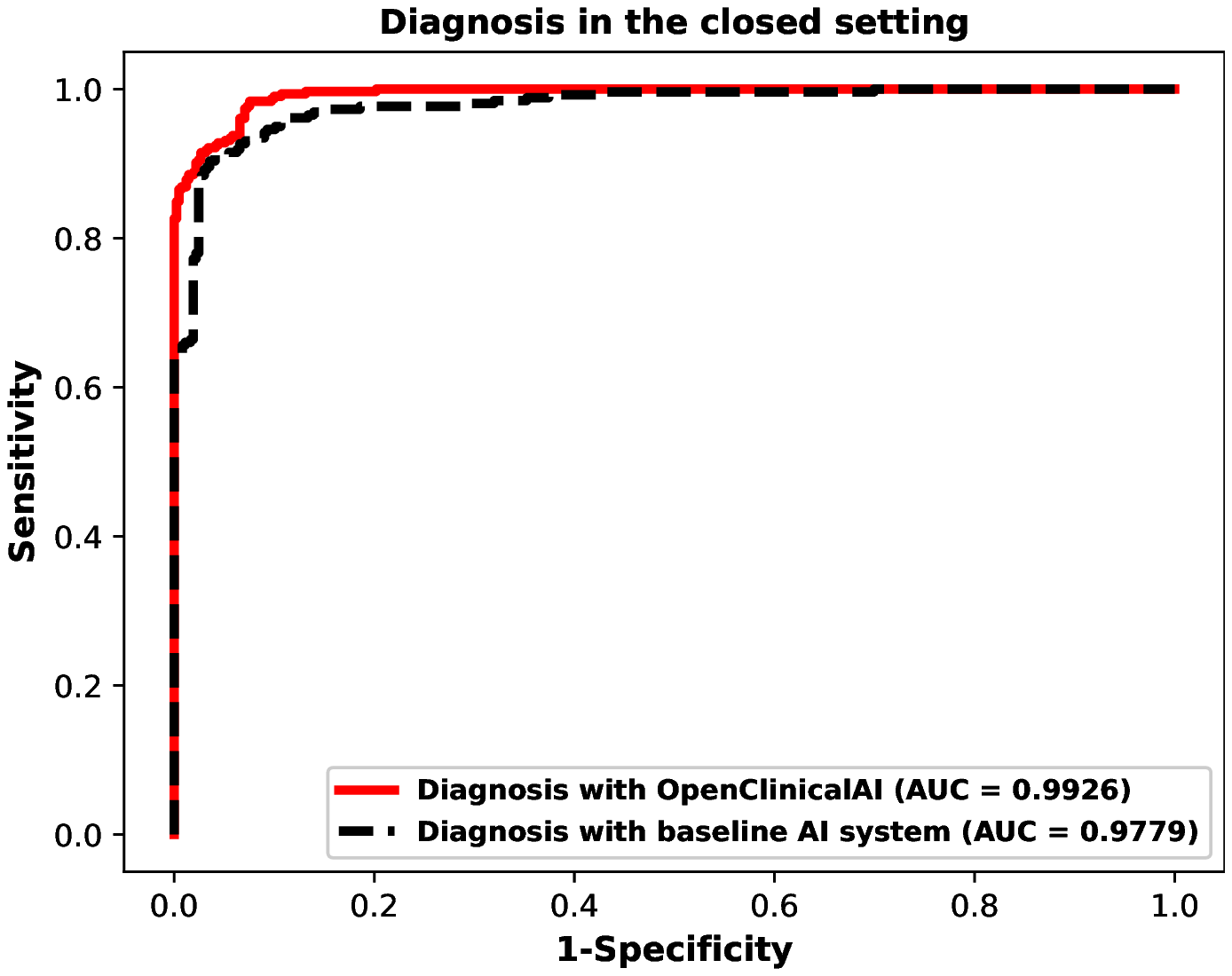} 
\end{minipage}
%\hspace{0.5cm}
\begin{minipage}[a]{0.48\textwidth}
\leftline{\textbf{b}}
\includegraphics[height=5.5cm, width=7.5cm]{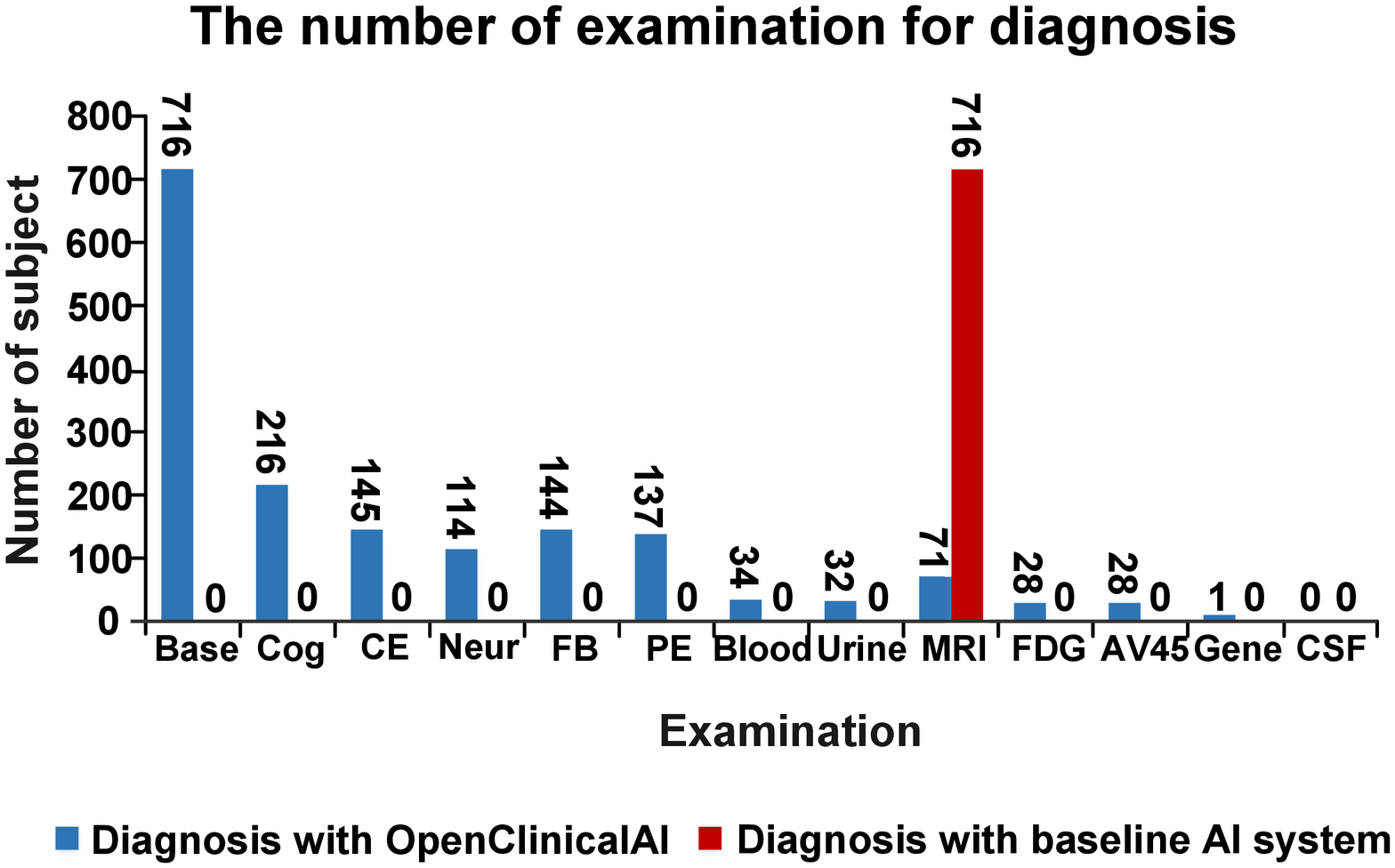} 
\end{minipage}\\
\quad
\caption{\textbf{The performance of OpenClinicalAI with personalized strategies against the baseline system on Alzheimer's disease diagnosis task in the closed  setting. a, } The ROC curves of two systems in the closed dataset. The red curve is the ROC curve of the baseline system, and it obtains an AUC score of  $0.9779$ (95\% CI 0.9722-0.9827). The black curve is the ROC curve of OpenClinicalAI with various examination data, and it obtains an AUC score of $0.9926$ (95\% CI 0.9907-0.9945). \textbf{b, } The examination used during the AD diagnosis process. The baseline system consistently uses MRI data and historical data as the system input. In other words, every subject must have a nuclear magnetic resonance scan. OpenClinicalAI is able to develop and adjust the diagnosis strategies according to individual conditions and existing medical conditions during the diagnosis process, and only $71$ subjects in the test set should have a nuclear magnetic resonance scan. Most subjects only need to have two or several simple examinations during the diagnosis process.\label{performance_closed}}
\end{figure*}

As shown in Fig.~\ref{performance_closed} a, the baseline system obtains a high AUC score of $0.9779$ (95\% confidence interval (CI) 0.9722-0.9827), and there is not much room for promotion. OpenClinicalAI achieves an AUC score of $0.9926$ (95\% CI 0.9907-0.9945) and obtains the state-of-the-art performance. However, the essential improvement from the baseline system to OpenClinicalAI is that the latter can dynamically develop personalized diagnosis strategies according to specific subjects and medical institutions. As shown in Fig.~\ref{performance_closed} b, less than 10\% of the subjects require a nuclear magnetic resonance scan, and most of the subjects only require harmless examination such as cognitive examination. We conclude OpenClinicalAI can avoid unnecessary examination for subjects and suit medical institutions with different examination abilities \footnote{Different hospitals have various clinical settings, such as community hospitals without nuclear magnetic resonance machines, big hospitals with multiple facilities.}.

\subsection*{The performance of OpenClincalAI against the baseline system in the real-world setting.}
Our goal is to develop an AI diagnosis system that can be embedded in the current medical system and cooperated with clinicians. In this work,  if the predicted probability of the AD or CN is smaller than the probability threshold ( 0.95 ), the subject will be marked as unknown and referral to the clinician. For comparison, we use the same baseline system discussed above. In addition, we also consider OpenClinicalAI without an OpenMax mechanism ( Algorithm S2,3) as the comparison system~\cite{bendale2015towards}.

\begin{figure*}
\centering
\begin{minipage}[a]{0.48\textwidth}
%\caption*{a.}
\leftline{\textbf{a}}
\includegraphics[height=4.8cm, width=7cm]{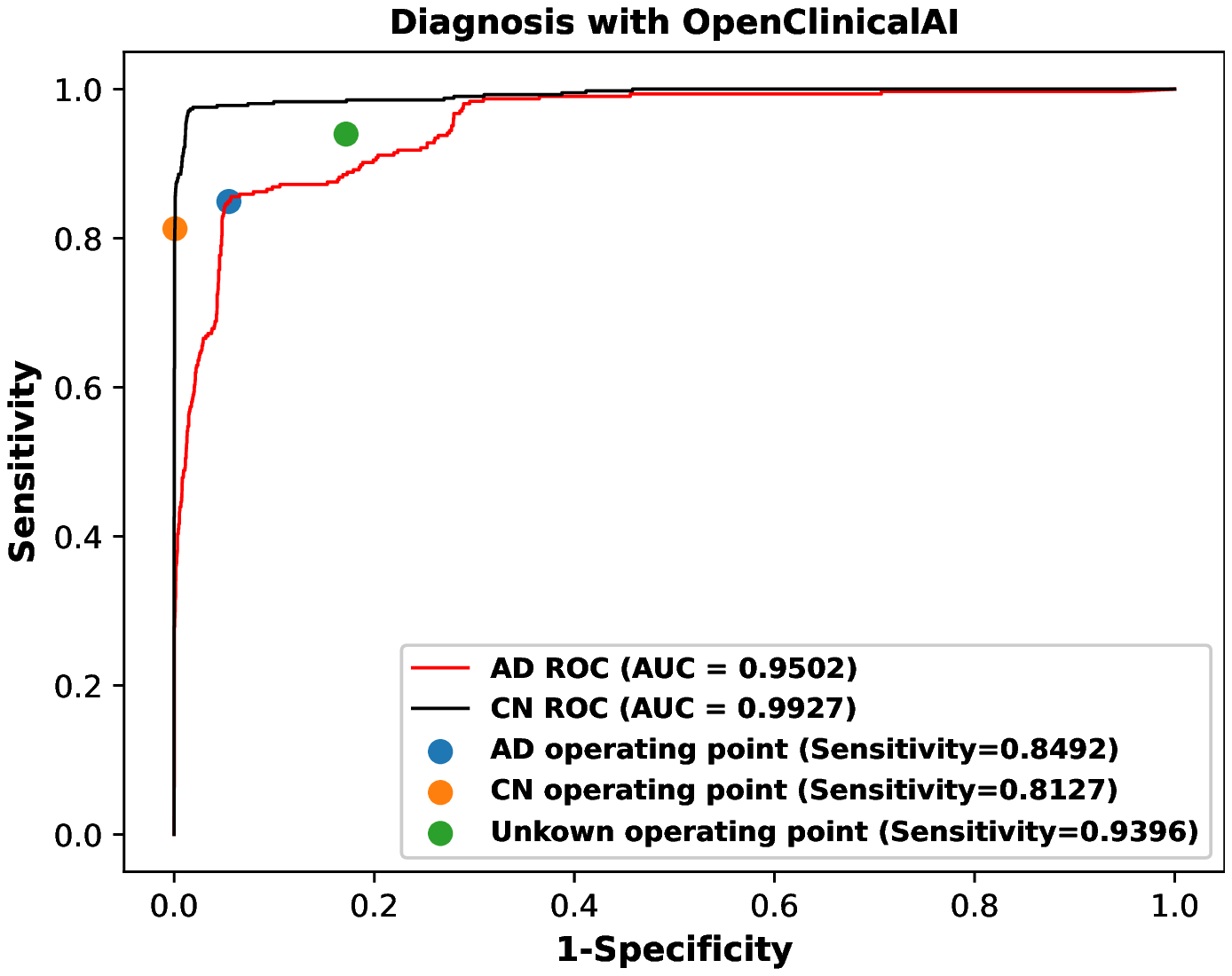} 
\end{minipage}
%\hspace{0.5cm}
\begin{minipage}[a]{0.48\textwidth}
\leftline{\textbf{b}}
\includegraphics[height=4.8cm, width=7cm]{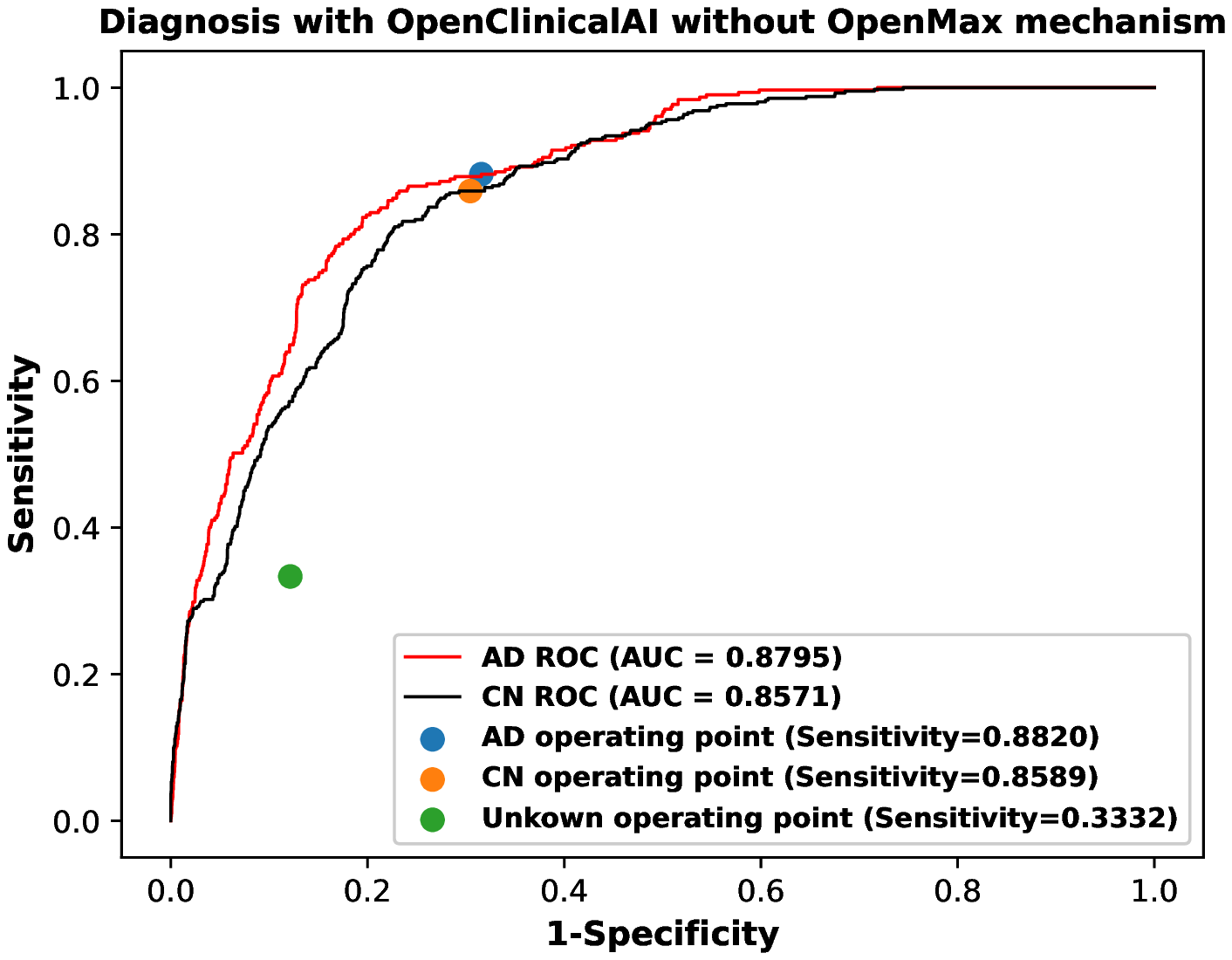} 
\end{minipage}\\
\quad
\begin{minipage}[a]{0.48\textwidth}
\leftline{\textbf{c}}
\includegraphics[height=4.8cm, width=7cm]{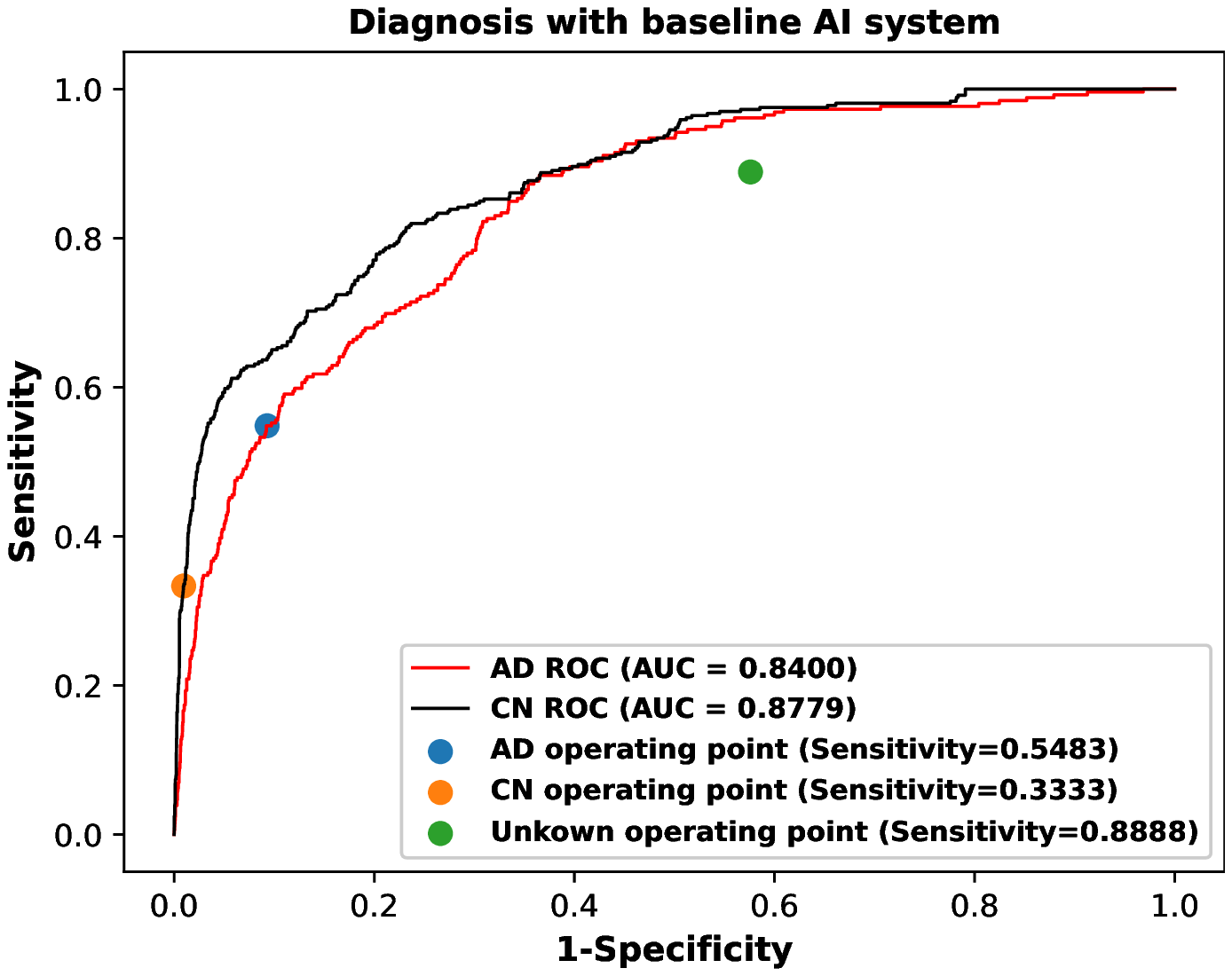} 
\end{minipage}
\begin{minipage}[a]{0.48\textwidth}
\leftline{\textbf{d}}
\includegraphics[height=4.8cm, width=7cm]{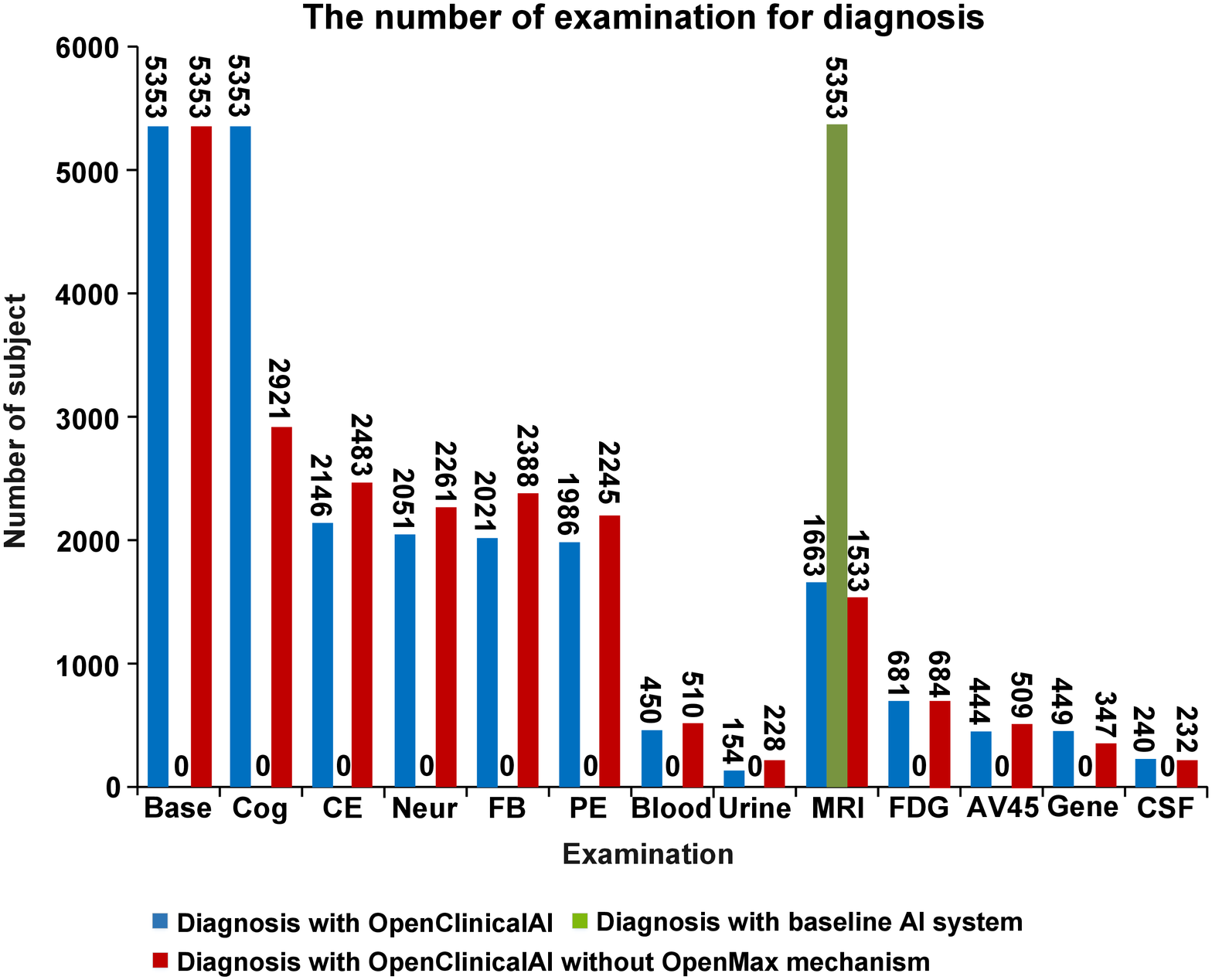} 
\end{minipage}\\
\quad
\caption{\textbf{The performance of OpenClinicalAI with personalized strategies against the baseline system in the real-world setting. a,} The ROC curves of OpenClinicalAI. It obtains two high AUC scores of 0.9502 (95\% CI 0.9304-0.9662) and 0.9927 (95\% CI 0.9854-0.9981) for AD and CN detection. The operating point of AI system is a group of score thresholds that separates positive and negative decisions of every category of the subject (0.95 for AD, 0.95 for CN, and 0.8 for unknown). On the operating point, OpenClinicalAI obtains the sensitivity of AD, CN, and unknown are $0.8492$ (95\% CI 0.7891-0.9051), $0.8127$ (95\% CI 0.7551-0.8667), and $0.9396$ (95\% CI 0.9290-0.9492) respectively. \textbf{b,} The ROC curves of OpenClinicalAI without an OpenMax mechanism. It obtains two AUC scores of 0.8795 (95\% CI 0.8540-0.9038) and 0.8571 (95\% CI 0.8331-0.8797) for AD and CN detection. On the operating point, the sensitivity of AD and CN are $0.8820$ (95\% CI 0.8282-0.9324) and $0.8589$ (95\% CI 0.8100-0.9056), respectively. However, the sensitivity of the unknown is only $0.3332$ (95\% CI 0.3133-0.3528). \textbf{c,} The ROC curves of the baseline system. The baseline system obtains two AUC scores of 0.8400 (95\% CI 0.8055-0.8728) and 0.8779 (95\% CI 0.8506-0.9025) for AD and CN detection. On the operating point, the unknown's sensitivity is $0.8888$ (95\% CI 0.8753-0.9018). However, the sensitivity of AD and CN are only $0.5483$ (95\% CI 0.4604-0.6301) and $0.3333$ (95\% CI 0.2663-0.3979), respectively. \textbf{d,}  The examination used during the AD diagnosis process. All subjects diagnosed by the baseline system require the nuclear magnetic resonance scan. The subject diagnosed by OpenClinicalAI without an OpenMax mechanism is similar to the subject diagnosed by OpenClinicalAI with an OpenMax mechanism. The selection of examination depends on the situation of the subject and existing medical conditions. Thus the examination number is not fixed.  \label{performance_open}}
\end{figure*}

As shown in Fig.~\ref{performance_open}a, b, and c, compared to the baseline  system, OpenClinicalAI demonstrates a significant improvement in the AUC of identification of AD subjects (+0.1102) and the AUC of identification of CN subjects (+0.1148). It is worth noting that OpenClinicalAI has a vast improvement in the sensitivity of AD, CN, and unknown on the operating point.

For the baseline system, the sensitivity of known (AD and CN) subjects is low. The sensitivity of AD is just 0.5483 (95\% CI 0.4604-0.6301), and the sensitivity of CN is just 0.3333(95\% CI 0.2663-0.3979). It indicates that most known subjects will be marked as unknown and sent to the clinician for diagnosis. Moreover, the sensitivity of unknown subjects is 0.8888(95\% CI 0.8753-0.9018), meaning 11.12\% of unknown subjects will be misdiagnosed. In addition, the baseline system requires that every subject has a nuclear magnetic resonance scan, and every medical institution that deploys the baseline system must be equipped with a nuclear magnetic resonance apparatus.

For OpenClinicalAI without an OpenMax mechanism, the sensitivity of known (AD and CN) subjects is as good as OpenClinicalAI with an OpenMax mechanism. In contrast, the sensitivity of unknown subjects is much worse than OpenClinicalAI with an OpenMax mechanism. It means most unknown subjects will be misdiagnosed, and it is unendurable in real-world settings. 

OpenClinicalAI diagnoses most of the known (AD and CN) subjects correctly, marks most of the rest as unknown, and sends them to the clinician for further diagnosis. Besides, most unknown subjects are correctly identified, and the misdiagnosis of unknown subjects is only $6.04\%$. It means that OpenClinicalAI has enormous potential application value to implement in real-world  settings. In addition, as shown in Fig.~\ref{performance_open}d, similar to the behaviors of OpenClinicalAI in the closed  setting, OpenClinicalAI can develop and adjust diagnosis strategies for every subject dynamically in the real-world setting. Only a small part of subjects require a nuclear magnetic resonance scan and more costs (economy and harm) examinations.

\subsection*{Development of diagnosis strategies}
For every subject, firstly, OpenClinicalAI will acquire the base information of the subject. Secondly, OpenClinicalAI will give a final diagnosis or receive other examination information according to the current data of the subject. Thirdly, repeat the previous step until the diagnosis is finalized or there is no further examination. 

\begin{figure*}
\centering
\begin{minipage}[a]{0.45\textwidth}
%\caption*{a.}
\leftline{\textbf{a}}
\includegraphics[height=6.24cm, width=7.21cm]{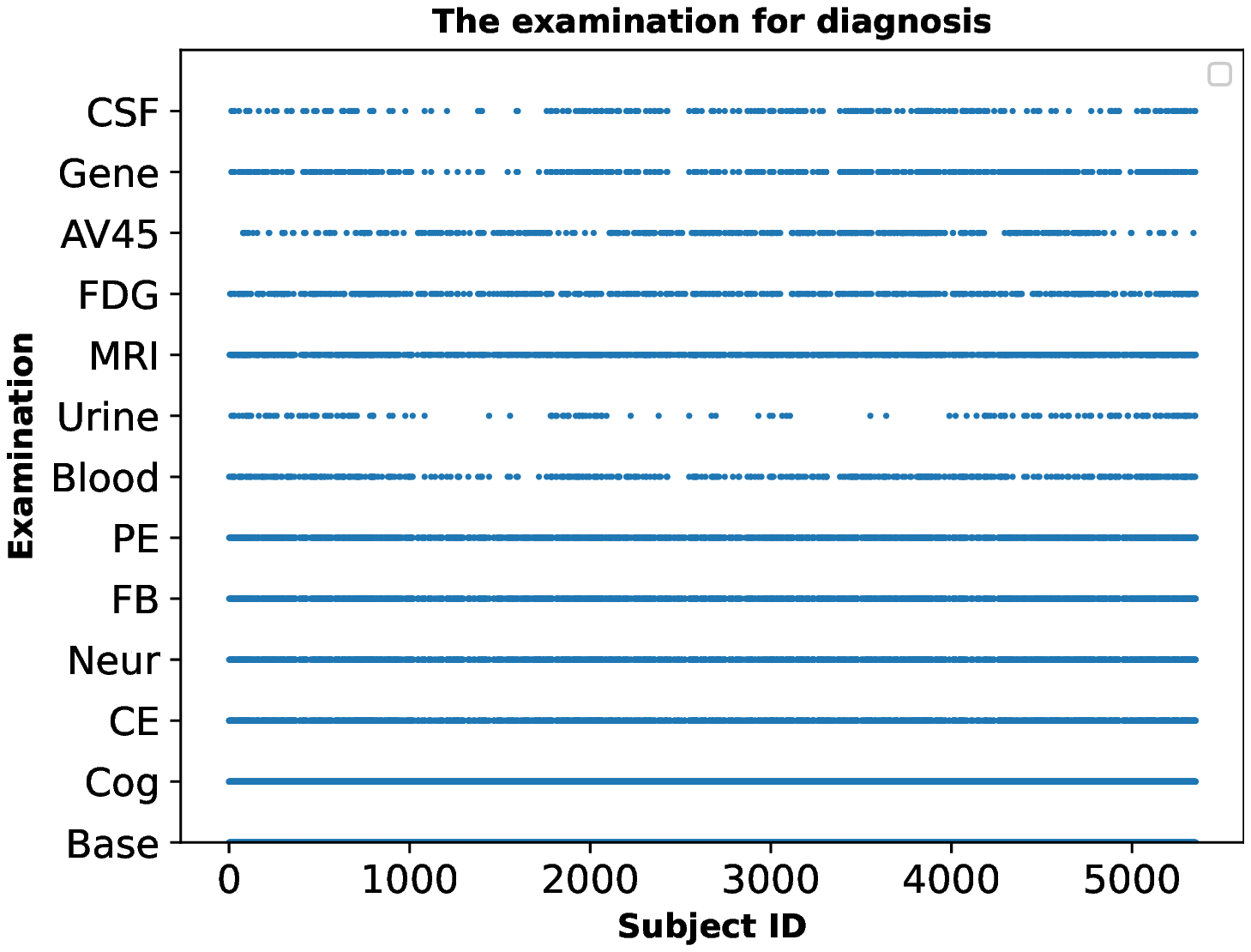} 
\end{minipage}
%\hspace{0.5cm}
\begin{minipage}[a]{0.45\textwidth}
\leftline{\textbf{b}}
\includegraphics[height=6.24cm, width=7.21cm]{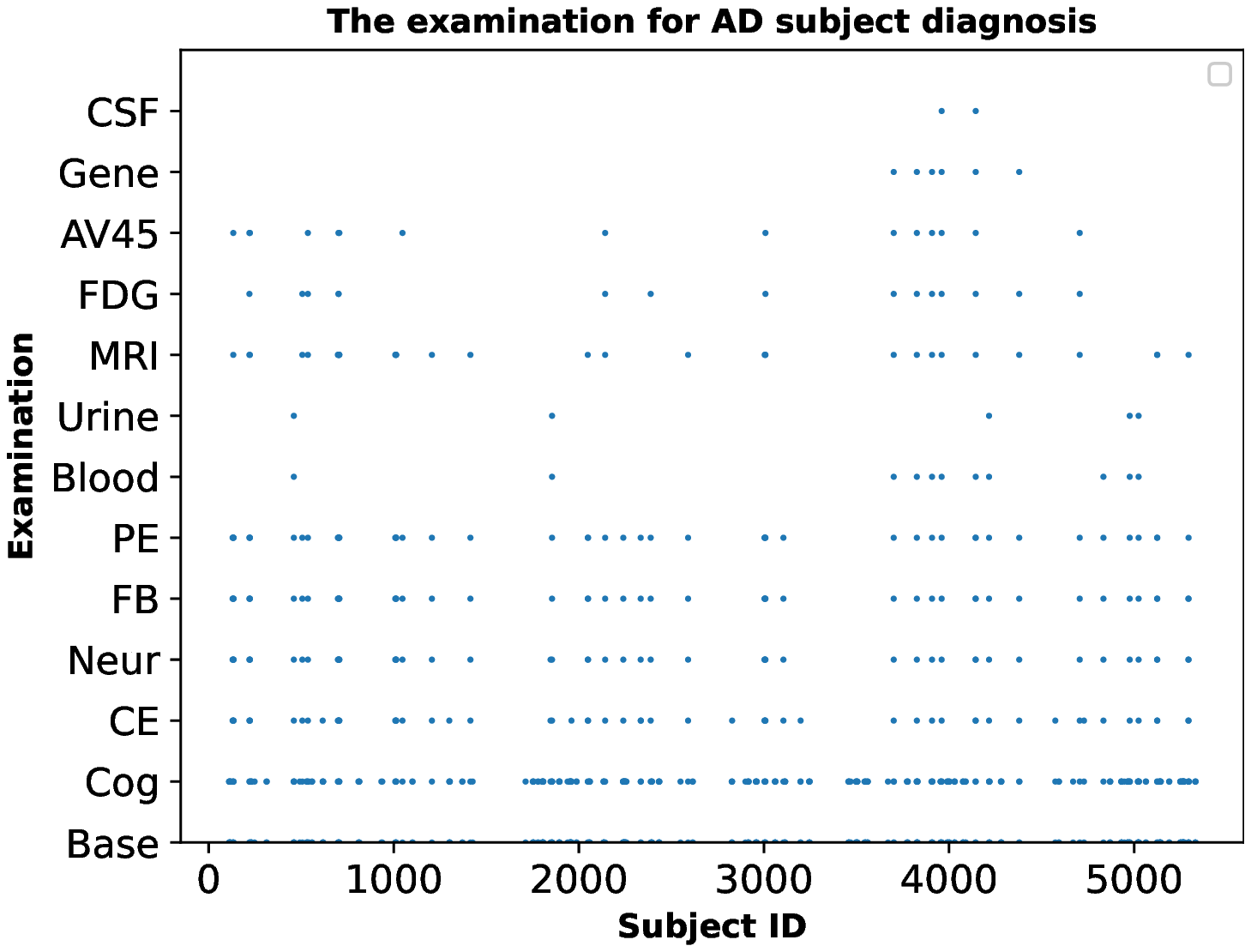} 
\end{minipage}\\
\quad
\begin{minipage}[a]{0.45\textwidth}
\leftline{\textbf{c}}
\includegraphics[height=6.24cm, width=7.21cm]{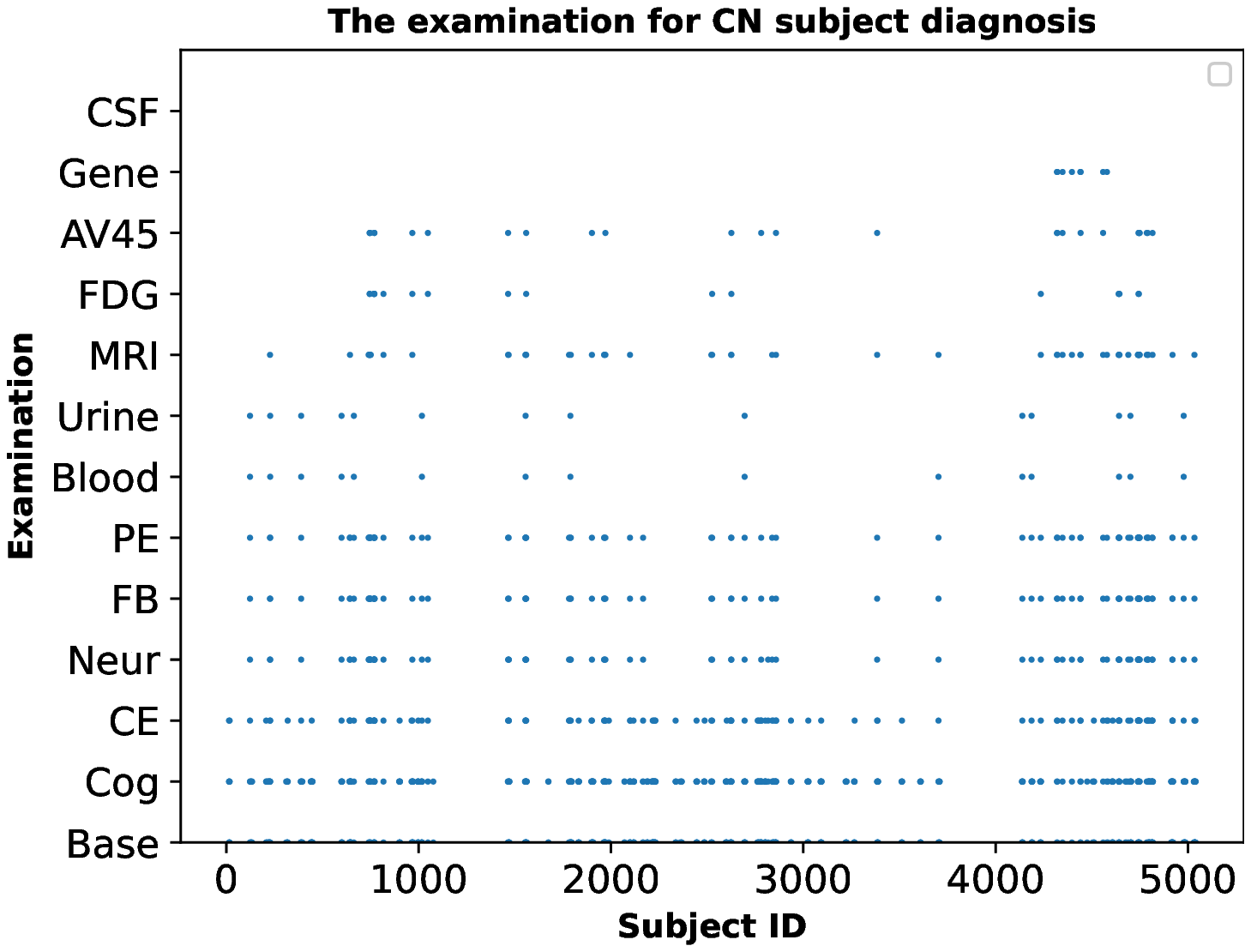} 
\end{minipage}
\begin{minipage}[a]{0.45\textwidth}
\leftline{\textbf{d}}
\includegraphics[height=6.24cm, width=7.21cm]{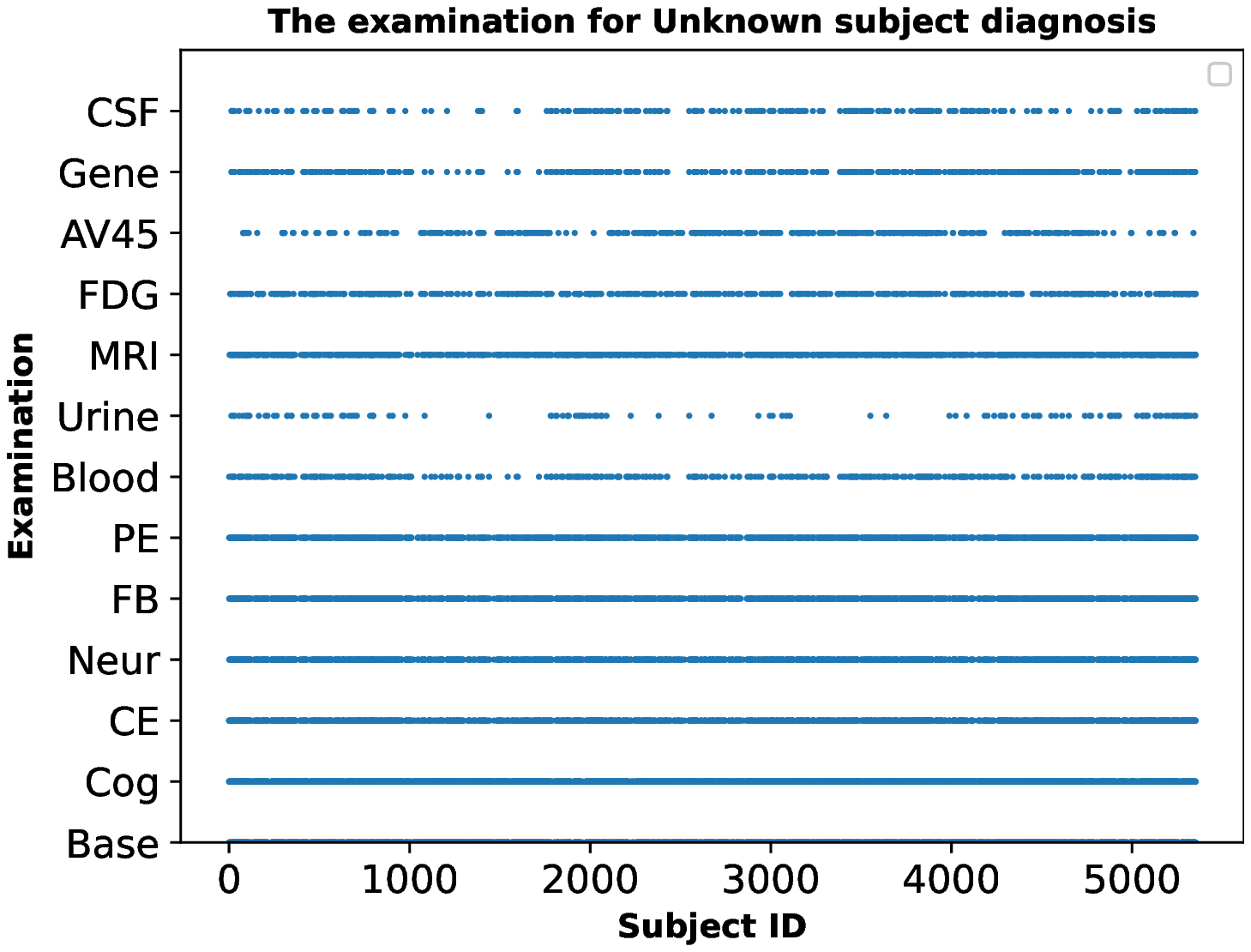} 
\end{minipage}\\
\quad
\caption{\textbf{Diagnosis strategies for subjects. a, } Diagnosis strategies for all subjects. Due to OpenClinicalAI developing and adjusting the examination for each subject, the selection of examinations for subjects is not the same. \textbf{b,} Diagnosis strategies for AD subjects. Compared to the high-cost examination, OpenClinicalAI pays more attention to the subject's basic information, cognitive, mental, behavioral, and physical examination information for the AD subject. In contrast, biochemical testing, imaging, and genetic data are less considered. \textbf{c,} Diagnosis strategies for CN subjects. The behaviors of OpenClinicalAI for CN recognition are similar to those for AD diagnosis, and the difference between those behaviors is that more examinations are required to identify the CN subject. \textbf{d,} Diagnosis strategies for unknown (MCI and SMC) subjects. Compared to the known subject recognition, identifying unknown subjects is more complicated, and more examinations are required. \label{strategy}}
\end{figure*}

As shown in Fig.~\ref{strategy}a, diagnosis strategies of subjects are not the same ( Table S3). OpenClinicalAI dynamically develops 35 diagnosis strategies according to different subject situations and all 40 examination abilities in the test set( Table S4). For the known (AD and CN) subjects, as shown in Fig.~\ref{strategy}b, and c, most of the subjects require low-cost examinations (such as cognition examination (CE)). A small part of subjects requires high-cost examinations (such as cerebral spinal fluid analysis (CSF) ). For unknown subjects, as shown in Fig.~\ref{strategy}d, different from the diagnosis of known (AD and CN) subjects, identifying unknown subjects is more complex and more dependent on high-cost examinations. The reason for the above phenomenon is that according to the mechanism of OpenClinicalAI, it will do its best to distinguish whether the subject belongs to the known categories. When it fails, OpenClinicalAI will mark the subject as unknown. It means that the unknown subject will undergo more examinations than the known subject. The details of the high-cost examinations requirement are as follows.
\begin{itemize}
\item[(1)]33.94\% of unknown subjects require a nuclear magnetic resonance scan (that of the known subject is 12.43\%).
\item[(2)]13.95\% of unknown subjects require a positron emission computed tomography scan with 18-FDG ( that of the known subject is 4.75\%). 
\item[(3)]8.67\% of unknown subjects require a positron emission computed tomography scan with AV45 ( that of the known subject is 5.87\%). 
\item[(4)]9.38\% of unknown subjects require a gene analysis ( that of the known subject is 1.96\%). 
\item[(5)]5.13\% of unknown subjects require a cerebral spinal fluid analysis (that of the known subject is 0.28\%).
\end{itemize}

\subsection*{Potential clinical applications}
OpenClinicalAI enables that the AD diagnosis system can be implemented in uncertain and complex clinical settings to reduce the workload of AD diagnosis and minimize the cost of subjects.

To identify the known (AD and CN) subject with high confidence, the operating point of OpenClinicalAI is running with a high decision threshold (0.95). For the test set, OpenClinicalAI achieved a accuracy value of 92.47\% (95\% CI 91.36\%-93.44\%), AD sensitivity value of 84.92\% (95\% CI 78.91\%-90.51\%), CN sensitivity value of 81.27\% (95\% CI 75.51\%-86.67\%) while retaining an unknown sensitivity value of 93.96\% (95\% CI 92.90\%-94.92\%). In addition, OpenClinicalAI can cooperate with the senior clinician to identify the known subject. In this work, 15.08\% (95\% CI 9.49\%-21.09\%) of AD subjects and 18.73\% (95\% CI 13.33\%-24.49\%) of CN subjects are marked as unknown and sent to senior clinicians to diagnose. The work pattern is significant for the undeveloped area, which is a promising way to connect developed areas and undeveloped areas to reduce the workload, improve the overall medical services, and promote medical equity.
To minimize the subject cost and maximize the subject benefit, OpenClinicalAI dynamically develops personalized diagnosis strategies for the subject according to the subject's situation and existing medical conditions. 

For the subject, OpenClinicalAI will judge whether it can finalize the subject's diagnosis according to the currently obtained information of subjects. If the current data of the subject is not enough to support OpenClinicalAI to make a diagnosis, it will recommend the most suitable further examination for the subject. It will mitigate the over-testing plight, minimize the subject cost, and maximize the subject benefits. For the test set, $35$ different diagnosis strategies are applied to the subject by OpenClinicalAI ( Table S3). The details of the high-cost examination are as follows.
\begin{itemize}
 \item[(1)]31.07\% of subjects require a nuclear magnetic resonance scan. 
 \item[(2)]12.72\% of subjects require a positron emission computed tomography scan with 18-FDG. 
 \item[(3)]8.29\% of subjects require a positron emission computed tomography scan with AV45.
 \item[(4)]8.39\% of subjects require a gene analysis. 
 \item[(5)]4.48\% of subjects require a cerebral spinal fluid analysis.
 \end{itemize}

For the medical institution, before the system recommends an examination for a subject, OpenClinicalAI will inquire whether the medical institution can execute the examination. Suppose the medical institution cannot perform the examination. In that case, OpenClinicalAI will recommend other examinations until the current information of the subject is enough to support it to make a diagnosis or until all common examinations have been suggested and the subject is marked as unknown. It enables that OpenClinicalAI is able to deploy in the different medical institutions with various examination abilities. In this work, OpenClinicalAI diagnoses subjects on 40 conditions of medical institutions ( Table S4). In addition, for the subject of the test set, due to lack of the information of recommended examinations (which may be equal to the medical institution not having the ability to execute the recommended examination), OpenClinicalAI adjusts the diagnostic strategies $14654$ times. 

\subsection*{Discussion}

Currently, the media overhype the AI assistance diagnosis system. However, it is far from being mature to be implemented in real-world clinical settings. Many clinicians are gradually losing faith in the medicine AI ~\cite{van2020artificial,weaver2021machine,brocklehurst2017computerised,chen2017machine,maddox2019questions,head2018patient}. Similar to the first trough of AI, the high expectation and unsatisfactory practical implementation of medical AI may severely hinder the development of medical AI. In addition, compared performances of state-of-the-art AI systems on stringent conditions and real-world settings, solving well the diagnosis task on stringent conditions is not much help to solve the diagnosis task in the real-world setting.
It is time to draw the attention from the pure algorithm research in closed settings to systematic study in real-world settings, focusing on the challenge of tackling the uncertainty and complexity of real-world settings. In this work, we propose an open, dynamic machine learning framework to make the AI diagnosis system can directly deal with the uncertainty and complexity in the real-world setting. Based on our framework, an AD diagnostic system demonstrates huge potentiality to implement in the real-world setting with different medical environments to reduce the workload of AD diagnosis and minimize the cost of the subject. 

Although many AI diagnostic systems have been proposed, how to embed these systems into the current health care system to improve the medical service remains an open issue~\cite{mckinney2020international,kuo2021application,schneider2021reflections,bullock2020mapping}. OpenClinicalAI provides a reasonable way to embed the AI system into the current health care system. OpenClinicalAI can collaborate with clinicians to improve the clinical service quality, especially the clinical service quality of undeveloped areas. On the one hand, OpenClinicalAI can directly deal with the diagnosis task in the uncertain and complex real-world setting. On the other hand, OpenClinicalAI can diagnose typical patients of known subjects, while sending those challenging or atypical patients of known subjects  to the clinicians for diagnosis. Although AI technology is different from traditional statistics, the model of the AI system still learns patterns from training data. For typical patients, the model is easy to understand patterns from patients, while it is challenging to learn patterns for atypical patients. Thus, every atypical and unknown patient is needed to treat by clinicians especially. In this work, most of the known subjects are diagnosed by OpenClinicalAI, and the rest are marked as unknown and sent to the senior clinician. 

Over-testing has always been a concern and has been exacerbated in current AI-based diagnostic systems~\cite{o2021global,o2018overtesting}. As samples, the systems proposed by Lu et al., Ding et al., and Liu et al. achieved state-of-the-art performance. At the same time, they required every subject to have a positron emission computed tomography scan, which is unnecessary for most subjects in real-work  settings~\cite{lu2018multiscale,ding2019deep,liu2018classification}. However, OpenClinicalAI enables AI systems able to develop personalized diagnosis strategies to avoid unnecessary testing. OpenClinicalAI provides a possible way that can effectively reduce over-testing under strict supervision.

Notably, the experiment of this work does not contain a comparison with clinicians. There are two main reasons. First, OpenClinicalAI obtains an AUC value of  0.9927 (95\% CI 0.9854-0.9981) in the closed  setting. It is very close to the ground truth and unnecessary compared to clinicians. Second, the diagnosis patterns in real-world settings aim to diagnose typical patients of known subjects (which is usually easier to diagnose) and send atypical patients of known subjects ( which are generally difficult to diagnose) and unknown subjects to clinicians. 
The task of OpenClinicalAI 
%has a good solution ($0.8492$ (95\% CI 0.7891-0.9051) AD subjects and $0.8127$ (95\% CI 0.7551-0.8667) CN subjects are correctly identified, 15.08\% (95\% CI 9.49\%-21.09\%) AD subjects and 18.73\% (95\% CI 13.33\%-24.49\%) CN subjects are sent to the clinician to diagnose), and 
is quite different from that one of clinicians. Unlike current AI-based diagnostic systems, OpenClinicalAI performs as a new part of the whole healthcare system instead of replacing the role of clinicians. Therefore, it is not necessary to compare OpenClinicalAI to clinicians.

Although OpenClinicalAI is promising to impact the future research of the diagnosis system, several limitations remain. First, the prospective clinical studies of diagnosis of Alzheimer's disease will be required to prove the effectiveness of our system. Second, the data of collection and processing are required to follow the standards of ADNI. 

%\biboptions{numbers,sort&compress}
%\bibliographystyle{Science}
%\biboptions{numbers,sort&compress}
%\bibliography{reference}

% Your references go at the end of the main text, and before the
% figures.  For this document we've used BibTeX, the .bib file
% scibib.bib, and the .bst file Science.bst.  The package scicite.sty
% was included to format the reference numbers according to *Science*
% style.

%BibTeX users: After compilation, comment out the following two lines and paste in
% the generated .bbl file. 

%\bibliography{scibib}

%\bibliographystyle{Science}

\section*{Acknowledgments}
We thank Weibo Pan and Fang Li for downloading the raw data sets from Alzheimer's Disease Neuroimaging Initiative. \textbf{Funding:} This work is supported by the Project of Guangxi Science and Technology (No. GuiKeAD20297004 to Y. H.) and the National Natural Science Foundation of China (No.61967002 to S. T.). Data collection and sharing for this project was funded by the Alzheimer's Disease Neuroimaging Initiative (ADNI) (National Institutes of Health Grant U01 AG024904) and DOD ADNI (Department of Defense award number W81XWH-12-2-0012). ADNI is funded by the National Institute on Aging, the National Institute of Biomedical Imaging and Bioengineering, and through generous contributions from the following: AbbVie, Alzheimer's Association; Alzheimer's Drug Discovery Foundation; Araclon Biotech; BioClinica, Inc.; Biogen; Bristol-Myers Squibb Company; CereSpir, Inc.; Cogstate; Eisai Inc.; Elan Pharmaceuticals, Inc.; Eli Lilly and Company; EuroImmun; F. Hoffmann-La Roche Ltd and its affiliated company Genentech, Inc.; Fujirebio; GE Healthcare; IXICO Ltd.;Janssen Alzheimer Immunotherapy Research \& Development, LLC.; Johnson \& Johnson Pharmaceutical Research \& Development LLC.; Lumosity; Lundbeck; Merck \& Co., Inc.;Meso Scale Diagnostics, LLC.; NeuroRx Research; Neurotrack Technologies; Novartis Pharmaceuticals Corporation; Pfizer Inc.; Piramal Imaging; Servier; Takeda Pharmaceutical Company; and Transition Therapeutics. The Canadian Institutes of Health Research is providing funds to support ADNI clinical sites in Canada. Private sector contributions are facilitated by the Foundation for the National Institutes of Health (\url{https://www.fnih.org/}). The grantee organization is the Northern California Institute for Research and Education, and the study is coordinated by the Alzheimer's Therapeutic Research Institute at the University of Southern California. ADNI data are disseminated by the Laboratory for Neuro Imaging at the University of Southern California. 
\textbf{Author contributions:}
Y.H. conceptualized the study, designed the models, wrote the codes, collected and analyzed the data, and wrote the manuscript. N.W., S.T., L.M, T.H., and Z.J. conceptualized the study and revised the manuscript. F.Z., G.K., X.M, X.G, and R,Z. collected and analyzed the data. Z.Z., and J.Z. directed the project and revised the manuscript. 
\textbf{Competing interests:}
The authors declare no competing financial interest. 
\textbf{Data and materials availability:}
The data from Alzheimer's Disease Neuroimaging Initiative was used under license for the current study. Applications for access to the dataset can be made at~\url{http://adni.loni.usc.edu/data-samples/access-data/}. All original code has been deposited at the website \href{https://www.benchcouncil.org/}{BenchCouncil} and is publicly available as of the date of
publication.

%Here you should list the contents of your Supplementary Materials -- below is an example. 
%You should include a list of Supplementary figures, Tables, and any references that appear only in the SM. 
%Note that the reference numbering continues from the main text to the SM.
% In the example below, Refs. 4-10 were cited only in the SM.     
\section*{Supplementary materials}
Materials and Methods\\
%Supplementary Text\\
Figs. S1 to S3\\
Tables S1 to S6\\
Algorithms S1 to S4\\
References \textit{(48-70)}

% For your review copy (i.e., the file you initially send in for
% evaluation), you can use the {figure} environment and the
% \includegraphics command to stream your figures into the text, placing
% all figures at the end.  For the final, revised manuscript for
% acceptance and production, however, PostScript or other graphics
% should not be streamed into your compliled file.  Instead, set
% captions as simple paragraphs (with a \noindent tag), setting them
% off from the rest of the text with a \clearpage as shown  below, and
% submit figures as separate files according to the Art Department's
% instructions.
\section*{Figures}

\renewcommand{\figurename}{Fig.}

\clearpage

\begin{center}
\textbf{Supplementary Materials for\\OpenClinicalAI: enabling AI to diagnose diseases in real-world clinical settings} 
\end{center}

\section*{Materials and Methods}
\subsection*{Human subjects}
Data used in the preparation of this article were obtained from the Alzheimer's Disease Neuroimaging Initiative (ADNI) database (\url{http://adni.loni.usc.edu}). The ADNI was launched in 2003 as a public-private partnership, led by Principal Investigator Michael W. Weiner,MD. For up-to-date information, see \url{http://www.adni-info.org.}

The data is collected from 67 sites in the United States and Canada ~\cite{petersen2010alzheimer,weiner2010alzheimer,weiner2015impact,weiner2017alzheimer}. The subject in the dataset aged between $54.4$ and $91.4$ at the first visit. The interval of the subject follow-up is usually greater than $6$ months. Generally, the longer the follow-up time is, the longer the interval is. The first visit is marked as bl, and the other visit is marked as m\underline{xx} according to the time (For example, the visit takes place six months after the first visit is marked as m06). Detailed characteristics of the subject are shown in Table S\ref{ptd},\ref{visit}.

\subsection*{Dataset}
The data contains study data, image data, genetic data compiled by ADNI between 2005 and 2019. Considering the commonly used examinations and the concerned examinations in AD diagnosis by the clinician, 13 categories of data are selected. 

\begin{itemize}
 \item[(1)]Base information, usually obtained through consultation, includes demographics, family history, medical history, symptoms. 
 \item[(2)]Cognition information, usually obtained through consultation and testing, includes Alzheimer's Disease Assessment Scale, Mini-Mental State Exam, Montreal Cognitive Assessment, Clinical Dementia Rating, Cognitive Change Index. 
 \item[(3)]Cognition testing, usually obtained through testing, includes ANART, Boston Naming Test, Category Fluency-Animals, Clock Drawing Test, Logical Memory-Immediate Recall, Logical Memory-Delayed Recall, Rey Auditory Verbal Learning Test, Trail Making Test. 
 \item[(4)]Neuropsychiatric information, usually obtained through consultation, includes Geriatric Depression Scale, Neuropsychiatric Inventory, Neuropsychiatric Inventory Questionnaire. 
 \item[(5)]Function and behavior information, usually obtained through consultation, includes Function Assessment Question, Everyday Cognitive Participant Self Report, Everyday Cognition Study Partner Report. 
 \item[(6)] Physical, neurological examination, usually obtained through testing, includes Physical Characteristics, Vitals, neurological examination. 
 \end{itemize}
 
The rest of the examinations include blood testing, urine testing, nuclear magnetic resonance scan, positron emission computed tomography scan with 18-FDG, positron emission computed tomography scan with AV45, gene analysis, and cerebral spinal fluid analysis. It is worth noting that not all categories of information are obtained for a subject's visit, and the information on each type is often incomplete. 

All subjects with labels containing at least one of the above categories of information are considered in this study. Two thousand one hundred twenty-seven subjects with 9593 visits are included in our work. A subject in a visit may require different categories of examination. Every combination of those examinations represents a diagnosis strategy. Thus, for the subject, $443795$ strategies are generated. These AD and CN subjects are randomly assigned to the training, validation, and test set. The training set contains 1025 subjects with 3986 visits and generates 180682 strategies. In the training set, 587 subjects with 1781 visits are AD and develop 80022 strategies, 466 subjects with 2205 visits are CN, and generate 100660 strategies. The validation set contains 73 subjects with 254 visits and generates 11898 strategies. In the validation set, 44 subjects with 127 visits are AD and develop 6008 strategies, 31 subjects with 127 visits are CN, and generate 5890 strategies. The test set contains 1460 subjects with 5353 visits. In the test set, 109 subjects with 305 visits are AD, 92 subjects with 411 visits are CN, 1082 subjects with 4357 visits are MCI, 280 subjects with 280 visits are SMC. Notably, the label of a subject may be different in other visits.

\subsection*{Randomization and blinding.}
AD and CN subjects as known categories of subjects are randomized into training, validation, and test sets by applying a random function provided by the Python3 tool. The assignment is determined by a float value generated by a random function. We assign subjects whose values are [0,0.8) into the training set, assign subjects whose values are [0.8,0.85) into the validation set, assign subjects whose values are [0.85,1] into the test set. The data of visits belong to the same subject are only allowed to appear in the same set.  MCI and SMC subjects as unknown categories of subjects are directly into the test set. During the development of the AI system, the test set is inaccessible. 
\subsection*{Data preparation.}
For each category of study data, if it contains more than one sub-category of data, concatenate all of the sub-category data by RID (The ID of the subject) and VISCODE (The mark of the subject's visit). For the medical image, we first convert the data from the DICOM format to the NIfNI format by the dcm2nii library. Second, register the image by ant library~\cite{5cf2edbbe75b4f90ac4c89e6ef1bc3f6,smith2004advances,senecaimproved}. Third, convert the 3D image to 2D slices and convert the image from gay to RGB. Finally, a trained model named DenseNet201 is used to extract the features of the 2D slices~\cite{huang2017densely}. For the genetic data, we extract 70 single nucleotide polymorphisms (SNP), which are very relating  to the AD ( Table S\ref{snp_ad}), and use one-hot code to represent each SNP~\cite{lambert2013meta,kunkle2019genetic,desikan2017genetic}. This work proposes a unified data representation framework, since the different dimensions of each category of data, the number of data categories included in each visit is different, and the number of history visits included in each subject is also different. We present an examination category in the subject's visit by an array with a shape of  $1\times2090$. The shape of our data is $n\times2090$, $n$ is the number of categories of data for the subject ( Fig. S\ref{Extent_fig2}).
\subsection*{The propose model}
Our model consists of five parts: $Encoder\_1$, $Decoder$, $Classifier\_1$, $Encoder\_2$, and $Classifier\_2$ ( Fig. S\ref{Extent_fig1} ). We name the model consisting of $Encoder\_1$, $Decoder$, and $Classifier\_1$ as $sub\_model\_1$, which can identify the subject from open clinical settings~\cite{geng2020recent,perera2020generative,bendale2015towards}. We name the model consisting of $Encoder\_2$, and $Classifier\_2$  as $sub\_model\_2$, which can dynamically develop and adjust the diagnosis strategy according to the situation of subjects and existing medical conditions.
\subsection*{Loss function}
The $sub\_model\_1$ is a multi-task learning model, which simultaneously optimizes the model's disease diagnosis and data reconstruction ability. The data reconstruction task can improve the diagnosis ability of the model in the open world~\cite{perera2020generative}. The loss function of the model is $Loss=0.65*loss_{diagnosis}+0.35*loss_{reconstruction}$. The $loss_{diagnosis}$ is categorical cross-entropy, and the $loss_{reconstruction}$ is mean squared logarithmic error. The $sub\_model\_2$ is also a multi-task learning model,  which simultaneously optimizes the 12 examinations whether should be selected as the next examination for the subject. We introduce a loss function that combines the BCE loss function and weighs losses with uncertainty~\cite{kendall2018multi,lee2019explainable}. The modified loss function is given by equations~\ref{eq1}:
%We introduced the ICH binary value (Fig. 3, M2) to update the output of the network to y = {y1, y2, y3, y4, y5, y6}, giving the probabilities of IPH, IVH, SDH, EDH, SAH and ICH, respectively. The multi-label classification task was then reformulated into a binary classification, with ICH as positive if one or more of the subtype outputs were positive and negative if not. The BCE loss function was weighted by the ratios of positive and negative instances for each class label, in a similar fashion as described previously11,12. The modified BCE loss function is given by equations (2) and (3):
\begin{equation}\label{eq1}\centering
\left.
  \begin{array}{c}
  Loss=-\sum\limits_{i=2}^{k}\frac{1}{2\delta_i^2}(\gamma_P^iy^ilog\hat{y}^i+\gamma_N^i(1-y^i)log(1-\hat{y}^i))+log\delta_i\\
  \\
  \gamma_P^i=\frac{|P^i|+|N^i|}{2|P^i|}  \qquad   \gamma_N^i=\frac{|P^i|+|N^i|}{2|N^i|}\\
  \\
\end{array}\right.
\end{equation}
where $|P^i|$ is the total number of $ith$ examinations as the subsequent examination, $|N^i|$ is the total number of other examinations as the following examination.  $\delta_i$ is an observation noise scalar of the output of $ith$ examination~\cite{kendall2018multi}.
\subsection*{Label examination}
Although researchers have made many efforts on the interpretability and internal logic of deep learning, the current behavior of deep learning is still tricky to understand~\cite{zhang2020effect,linardatos2021explainable}. We do not know whether the diagnosis strategy of the AI model needs to be consistent with human experts. Thus, it is unnecessary to label the subsequent examination of the current examination strategy by the clinician and train a model to simulate the clinician's behavior. In this work, the following examination label is labeled by the examination label algorithm ( Algorithm S\ref{alg1} ). 
The subsequent examination for the subject is determined by whether this examination makes the prediction model ($sub\_model\_1$) obtain a greater predicted probability for the correct category and smaller predicted probabilities for other categories. 
\subsection*{OpenMax}
OpenMax is a modified SoftMax layer that adopted the concept of Meta-Recognition~\cite{ge2017generative,bendale2015towards,Scheirer_2011_TPAMI}. OpenMax uses the distance between the activation vector (AV) of the sample and the mean activation vector (the mean computed over only the correctly classified training examples) to identify the unknown categories of the subject ~\cite{bendale2015towards}. The deep learning network can be regarded as a feature extractor, and the output of the AV layer can be regarded as characteristics of the sample. However, the AV layer usually only retains the most relevant features to the classification task, and the features related to the unknown category are not guaranteed to be retained. To alleviate this problem, we replaced the output of the AV layer with the abnormal patterns of 14 selected indicators of known categories according to the Alzheimer's Diagnosis guidelines to improve the performance of the AI model~\cite{jack2011introduction,sperling2011toward,albert2011diagnosis,donohue2014preclinical} ( Table S\ref{index_nomal}  ). The modified OpenMax by abnormal patterns is shown in Algorithm S\ref{alg2},\ref{alg3}. 
\subsection*{Model training.}
The training of our model consists of two stages. The first stage is training the $sub\_model\_1$, in which the $Classifier\_1$ uses SoftMax layer as the output layer. The dimension of the output of the $sub\_model\_1$ in this training stage is 2, corresponding to AD and CN. After training the $sub\_model\_1$, a modified OpenMax layer, which estimates the probability of an input being an unknown class, is used to replace the SoftMax layer~\cite{bendale2015towards}. The dimension of the output of $sub\_model\_1$ in the prediction stage is 3, corresponding to AD, CN, and unknown. According to prediction probabilities of subjects by the $sub\_model\_1$, every examination strategy in the training set and validation set is labeled by the Algorithm S\ref{alg1}. The second stage is training the $sub\_model\_2$, the input of the $sub\_model\_2$ contains raw data and the prediction probability, the dimension of the output of the $sub\_model\_2$ is 12, which respectively correspond to 12 categories of examination.
The model was optimized using mini-batch stochastic gradient descent with Adam and a base learning rate of 0.0005~\cite{kingma2015adam}. The experiments are conducted on a Linux server equipped with
Tesla P40 and Tesla P100 GPU.

Due to the historical information has a significant influence on the diagnosis of Alzheimer's disease, there is a vast difference between the diagnosis of Alzheimer's disease at first visit without historical information and other visits with historical data. Therefore, based on the above model training method, we additionally trained a model for diagnosing Alzheimer's disease at the first visit based on the subject's data at the first visit.
\subsection*{Prediction}
Unlike the other state-of-the-art AI models, predictions of our model are dynamic. The prediction algorithm comprehensively considers the situation of the subject, the condition of the medical institution, and the ability of our model to dynamically adjust the diagnosis strategy ( Algorithm S\ref{alg_2}). Firstly, our model will generate the probability for every category (AD vs. CN vs. Unknown) according to the current input data of the subject. Second, if the probability of categories exceeds thresholds (AD $0.95$, CN $0.95$, unknown $0.8$), output the corresponding label. Otherwise, adjust the examination strategy by selecting the subsequent examination according to the situation of the subject and the medical institution, and go to the first step. Finally, if all diagnostic strategies are tried, the model still cannot obtain the probability of exceeding the threshold and then outputs unknown.
\subsection*{Statistical analysis}
To evaluate the evaluation index of the AI model, a non-parametric bootstrap method is applied to calculate the confidence intervals (CI)  for the evaluation index~\cite{efron1994introduction}. In this work, we calculate 95\% CI for every evaluation index. We randomly sample $2500$ cases from the test set and evaluated the AI model by the sampled set for every evaluation index. $2000$ repeated trials are executed, and $2000$
values of the evaluation index are generated. The 95\% CI is obtained by the 2.5 and 97.5 percentiles of the distribution of the evaluation index values.

%\biboptions{numbers,sort&compress}

\renewcommand{\figurename}{Fig. S}
\setcounter{figure}{0}
%\renewcommand{\thefigure}{\ifnum \c@chapter>\z@ \thechapter-\fi \@arabic\c@figure}
%\noindent\textbf{Supplemental information}
\begin{figure*}
\centering
\includegraphics[height=13cm, width=15cm]{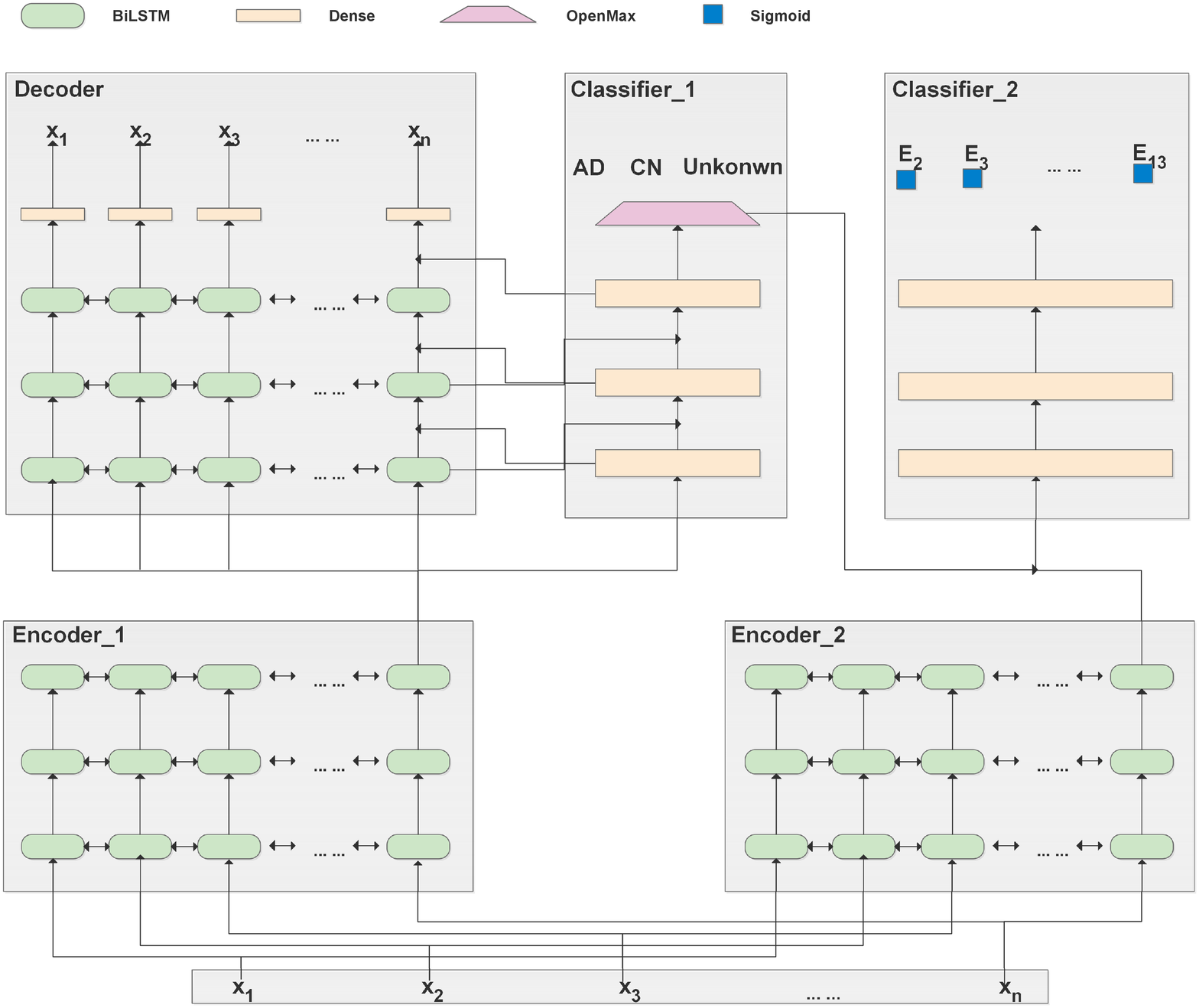} 
\caption{\textbf{The open, dynamic machine learning framework of OpenClinicalAI.} The OpenClinicalAI framework contains four independent modules and one accessory module. $Encoder\_1$ processes the input data for the $Classifier\_1$, and $Encoder\_2$ processes the input data for the $Classifier\_2$. The $Classifier\_1$ introduces the OpenMax mechanism to identify unknown categories of subjects. The accessory module $Decoder$ is used to help the $Classifier\_1$ retain features of the sample and improve the ability to identify unknown categories of subjects. The $Classifier\_2$ is used to select the examination to be carried out in the next step. In addition, the length of input data of the OpenClinicalAI framework is variable to adapt to data of different subjects at different visits. \label{Extent_fig1}}
\end{figure*}

\clearpage

\begin{figure*}
\centering
\includegraphics[height=12cm, width=13cm]{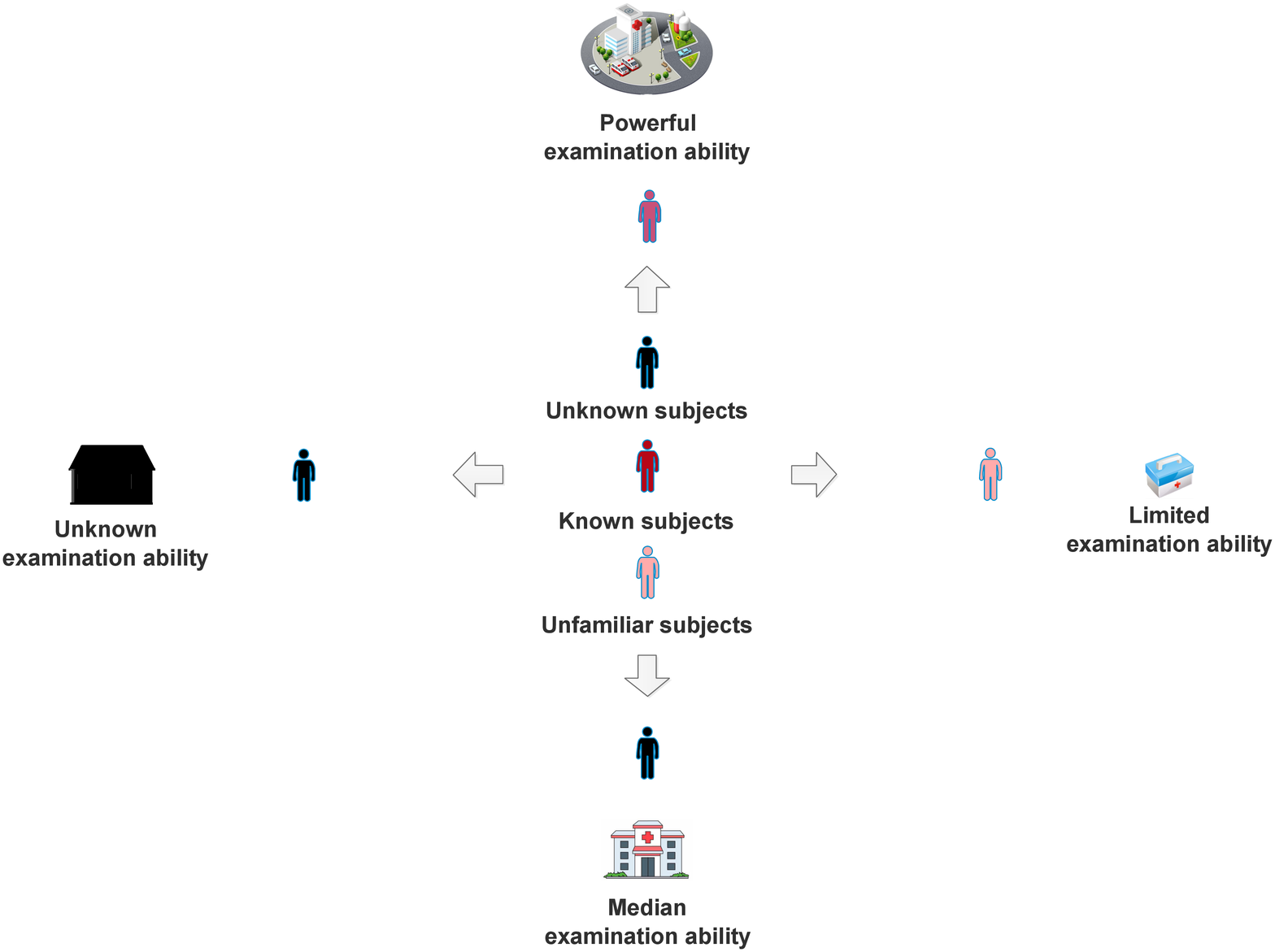} 
\caption{\textbf{The real-world setting of Clinical AIBench.}  Subjects in real-world settings are different with various situations. They contain different pre-known categories and unknown and unfamiliar categories for the specific clinician or AI diagnostic system. The visit of subjects to a particular medical institution can not be pre-specified and hence are uncertain. Medical institutions in real-world  settings also are different with different executive abilities of the examination. The executive ability of the examination in various medical institutions is very different from small-scale country clinics to large-scale hospitals. In addition, it is difficult to know by advance all the specific medical institutions that will deploy the AI system and their particular executive abilities of the examination. \label{open_setting}}
\end{figure*}

\clearpage

\begin{figure*}
\centering
\includegraphics[height=9.13cm, width=16cm]{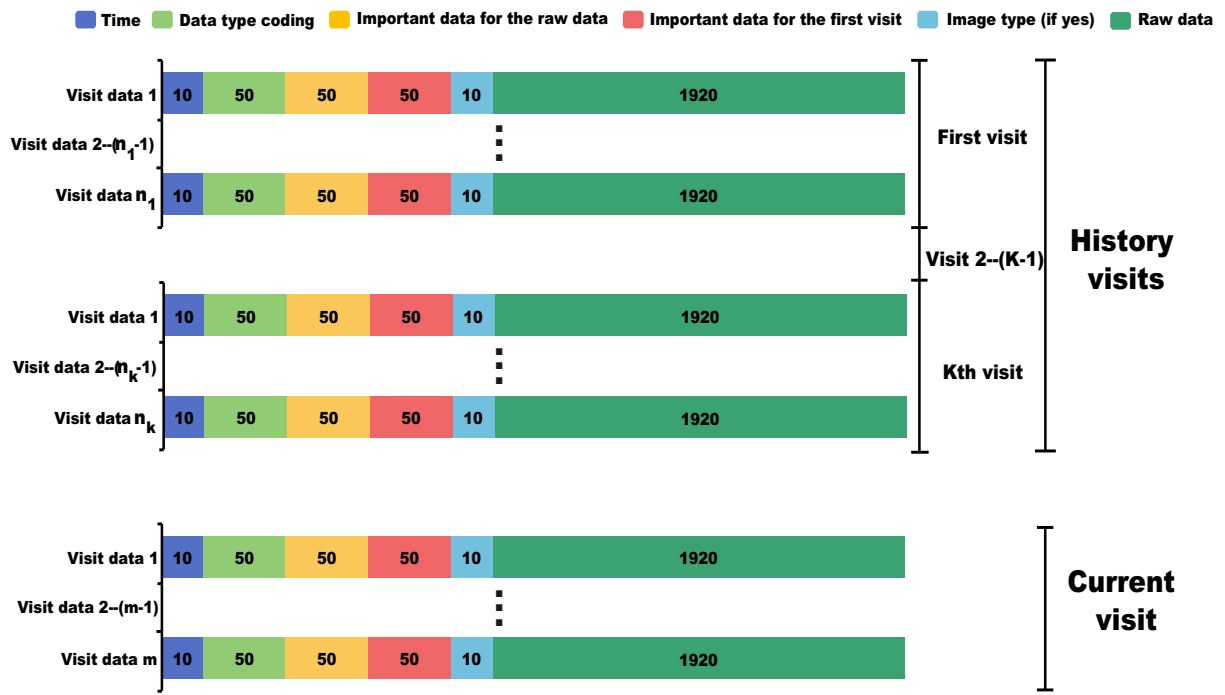} 
\caption{\textbf{Data framework for single subject.} Our data representation framework comprehensively considers the historical visit information and current visit information of the subject. The data with the earlier time is farther away from the current data. \label{Extent_fig2}}
\end{figure*}

% \usepackage{multirow}
%\begin{table*}[]

\clearpage

\renewcommand{\tablename}{Table S}
\setcounter{table}{0}

\begin{table}
\centering
\caption{\textbf{Characteristics of subjects.}\label{ptd}}
\begin{tabular}{cccccc}
\hline
                                                                            &                   & Data set & Training set & Validation set & Test set \\
\hline
\multirow{5}{*}{Age}                                                        & 54-59.9           & 80       & 36           & 2              & 59       \\
                                                                            & 60-69.9           & 596      & 246          & 10             & 442      \\
                                                                            & 70-70.9           & 1048     & 528          & 46             & 695      \\
                                                                            & 80-80.9           & 395      & 213          & 14             & 259      \\
                                                                            & 90-91.9           & 6        & 1            & 1              & 4        \\
\multirow{2}{*}{Gender}                                                     & Female            & 1130     & 560          & 44             & 785      \\
                                                                            & Male              & 997      & 465          & 29             & 675      \\
\multirow{5}{*}{Educate}                                                    & 4-7               & 11       & 4            & 0              & 8        \\
                                                                            & 8-10              & 40       & 18           & 2              & 23       \\
                                                                            & 11-13             & 353      & 176          & 13             & 243      \\
                                                                            & 14-16             & 823      & 403          & 26             & 558      \\
                                                                            & 17-20             & 900      & 424          & 32             & 628      \\
\multirow{3}{*}{\begin{tabular}[c]{@{}c@{}}Ethnic \\ category\end{tabular}} & Hisp/Latino       & 73       & 32           & 5              & 49       \\
                                                                            & Not Hisp/Latino   & 2042     & 986          & 67             & 1404     \\
                                                                            & Unknown           & 12       & 7            & 1              & 7        \\
\multirow{7}{*}{\begin{tabular}[c]{@{}c@{}}Racial \\ category\end{tabular}} & Asian             & 40       & 20           & 0              & 25       \\
                                                                            & Black             & 88       & 41           & 5              & 57       \\
                                                                            & Hawaiian/Other PI & 2        & 0            & 0              & 2        \\
                                                                            & More than one     & 25       & 10           & 0              & 18       \\
                                                                            & White             & 1964     & 954          & 68             & 1350     \\
                                                                            & Am Indian/Alaskan & 4        & 0            & 0              & 4        \\
                                                                            & Unknown           & 4        & 0            & 0              & 4        \\
\multirow{5}{*}{Marriage}                                                   & Married           & 1618     & 805          & 59             & 1100     \\
                                                                            & Never\_married    & 73       & 30           & 3              & 48       \\
                                                                            & Widowed           & 238      & 114          & 8              & 165      \\
                                                                            & Divorced          & 191      & 75           & 3              & 141      \\
                                                                            & Unknown           & 7        & 1            & 0              & 6        \\
\multicolumn{1}{l}{\multirow{4}{*}{Category}}                               & AD                & 740      & 587          & 44             & 109      \\
\multicolumn{1}{l}{}                                                        & CN                & 589      & 466          & 31             & 92       \\
\multicolumn{1}{l}{}                                                        & MCI               & 1082     & 0            & 0              & 1082     \\
\multicolumn{1}{l}{}                                                        & SMC               & 280      & 0            & 0              & 280   \\
\hline  
\end{tabular}
\end{table}

\clearpage

\begin{table}
\centering
\caption{\textbf{The visit distribution of subjects.}\label{visit}}
\begin{tabular}{ccccc}
\hline
Visit & Data set & Training set & Validation set & Test set \\
\hline
first visit    & 2126     & 705          & 53             & 1368     \\
m06    & 1515     & 604          & 45             & 866      \\
m12   & 1475     & 621          & 43             & 811      \\
m18   & 329      & 79           & 5              & 245      \\
m24   & 1217     & 591          & 32             & 594      \\
m36   & 804      & 338          & 23             & 443      \\
m48   & 638      & 311          & 14             & 313      \\
m60   & 399      & 178          & 13             & 208      \\
m72   & 395      & 207          & 14             & 174      \\
m84   & 268      & 128          & 6              & 134      \\
m96   & 146      & 74           & 3              & 69       \\
m108  & 100      & 55           & 1              & 44       \\
m120  & 75       & 39           & 1              & 35       \\
m132  & 55       & 33           & 0              & 22       \\
m144  & 39       & 19           & 1              & 19       \\
m156  & 12       & 4            & 0              & 8       \\
\hline
\end{tabular}
\end{table}

\clearpage

% \usepackage{multirow}
\begin{table}
\centering
\tiny
\caption{\textbf{The diagnosis strategies for the test set.}\label{methond}}

\begin{tabular}{cccccccccccccccccc}
\hline
&\multicolumn{13}{c}{Diagnosis strategies}                                          & \multicolumn{4}{c}{Visit number of subject} \\
&Base & Cog & CE & Neur & FB & PE & Blood & Urine & MRI & FDG & AV45 & Gene & CSF & AD      & CN      & Unknown     & Total     \\
\hline
\textbf{1}&1    & 1   & 0  & 0    & 0  & 0  & 0     & 0     & 0   & 0   & 0    & 0    & 0   & 244     & 280     & 2680        & 3204      \\
\textbf{2}&1    & 1   & 1  & 1    & 1  & 1  & 0     & 0     & 1   & 0   & 0    & 0    & 0   & 16      & 23      & 697         & 736       \\
\textbf{3}&1    & 1   & 1  & 1    & 1  & 1  & 0     & 0     & 1   & 0   & 1    & 0    & 0   & 3       & 10      & 81          & 94        \\
\textbf{4}&1    & 1   & 1  & 1    & 1  & 1  & 0     & 0     & 1   & 1   & 1    & 0    & 0   & 6       & 8       & 124         & 138       \\
\textbf{5}&1    & 1   & 1  & 0    & 0  & 0  & 0     & 0     & 0   & 0   & 0    & 0    & 0   & 8       & 47      & 43          & 98        \\
\textbf{6}&1    & 1   & 1  & 1    & 1  & 1  & 0     & 0     & 0   & 0   & 0    & 0    & 0   & 10      & 6       & 183         & 199       \\
\textbf{7}&1    & 1   & 1  & 1    & 0  & 0  & 0     & 0     & 0   & 0   & 0    & 0    & 0   & 1       & 2       & 27          & 30        \\
\textbf{8}&1    & 1   & 1  & 1    & 1  & 1  & 0     & 0     & 0   & 1   & 0    & 0    & 0   & 1       & 2       & 11          & 14        \\
\textbf{9}&1    & 1   & 1  & 1    & 1  & 1  & 0     & 0     & 0   & 0   & 1    & 0    & 0   & 1       & 1       & 16          & 18        \\
\textbf{10}&1    & 1   & 1  & 1    & 1  & 1  & 0     & 0     & 1   & 1   & 0    & 0    & 0   & 2       & 6       & 232         & 240       \\
\textbf{11}&1    & 1   & 1  & 1    & 1  & 1  & 0     & 0     & 0   & 1   & 1    & 0    & 0   & 0       & 3       & 16          & 19        \\
\textbf{12}&1    & 1   & 1  & 1    & 1  & 0  & 0     & 0     & 0   & 1   & 1    & 0    & 0   & 0       & 0       & 1           & 1         \\
\textbf{13}&1    & 1   & 1  & 1    & 1  & 0  & 0     & 0     & 0   & 0   & 0    & 0    & 0   & 1       & 0       & 33          & 34        \\
\textbf{14}&1    & 1   & 0  & 1    & 1  & 1  & 0     & 0     & 1   & 1   & 1    & 0    & 0   & 0       & 0       & 2           & 2         \\
\textbf{15}&1    & 1   & 0  & 1    & 1  & 1  & 0     & 0     & 1   & 0   & 0    & 0    & 0   & 0       & 0       & 1           & 1         \\
\textbf{16}&1    & 1   & 1  & 1    & 1  & 1  & 1     & 0     & 1   & 1   & 1    & 1    & 1   & 2       & 0       & 113         & 115       \\
\textbf{17}&1    & 1   & 1  & 1    & 1  & 1  & 1     & 1     & 0   & 0   & 0    & 0    & 0   & 5       & 14      & 10          & 29        \\
\textbf{18}&1    & 1   & 1  & 1    & 1  & 1  & 0     & 0     & 1   & 0   & 0    & 1    & 0   & 0       & 3       & 21          & 24        \\
\textbf{19}&1    & 1   & 1  & 1    & 1  & 1  & 1     & 0     & 1   & 1   & 1    & 1    & 0   & 3       & 0       & 13          & 16        \\
\textbf{20}&1    & 1   & 1  & 1    & 1  & 1  & 1     & 0     & 0   & 0   & 0    & 0    & 0   & 1       & 0       & 42          & 43        \\
\textbf{21}&1    & 1   & 1  & 1    & 1  & 1  & 1     & 0     & 1   & 0   & 0    & 0    & 0   & 0       & 1       & 2           & 3         \\
\textbf{22}&1    & 1   & 1  & 1    & 1  & 1  & 0     & 0     & 1   & 0   & 1    & 1    & 0   & 0       & 5       & 21          & 26        \\
\textbf{23}&1    & 1   & 1  & 1    & 1  & 1  & 0     & 0     & 1   & 1   & 0    & 1    & 0   & 1       & 0       & 9           & 10        \\
\textbf{24}&1    & 1   & 1  & 1    & 1  & 1  & 1     & 1     & 1   & 1   & 0    & 1    & 1   & 0       & 0       & 54          & 54        \\
\textbf{25}&1    & 1   & 1  & 1    & 1  & 1  & 1     & 0     & 1   & 0   & 0    & 1    & 0   & 0       & 0       & 60          & 60        \\
\textbf{26}&1    & 1   & 1  & 1    & 1  & 1  & 1     & 1     & 1   & 0   & 0    & 1    & 1   & 0       & 0       & 67          & 67        \\
\textbf{27}&1    & 1   & 1  & 1    & 1  & 1  & 1     & 0     & 1   & 1   & 0    & 1    & 0   & 0       & 0       & 55          & 55        \\
\textbf{28}&1    & 1   & 1  & 1    & 1  & 1  & 1     & 1     & 1   & 1   & 0    & 1    & 0   & 0       & 0       & 1           & 1         \\
\textbf{29}&1    & 1   & 1  & 1    & 1  & 1  & 1     & 1     & 1   & 0   & 0    & 1    & 0   & 0       & 0       & 3           & 3         \\
\textbf{30}&1    & 1   & 1  & 1    & 1  & 1  & 1     & 0     & 1   & 1   & 0    & 1    & 1   & 0       & 0       & 2           & 2         \\
\textbf{31}&1    & 1   & 1  & 1    & 1  & 1  & 0     & 0     & 1   & 1   & 1    & 1    & 0   & 0       & 0       & 13          & 13        \\
\textbf{32}&1    & 1   & 1  & 1    & 1  & 1  & 1     & 0     & 1   & 1   & 1    & 0    & 0   & 0       & 0       & 1           & 1         \\
\textbf{33}&1    & 1   & 1  & 1    & 1  & 1  & 1     & 0     & 1   & 0   & 0    & 1    & 1   & 0       & 0       & 1           & 1         \\
\textbf{34}&1    & 1   & 1  & 1    & 1  & 1  & 0     & 0     & 1   & 0   & 0    & 1    & 1   & 0       & 0       & 1           & 1         \\
\textbf{35}&1    & 1   & 1  & 1    & 1  & 1  & 0     & 0     & 0   & 0   & 1    & 1    & 0   & 0       & 0       & 1           & 1        \\
\hline
\end{tabular}
\end{table}

\clearpage

\begin{table}
\centering
\tiny
\caption{\textbf{Medical institutions with different examination abilities in the test set.} \label{institutions}}
\begin{threeparttable}
\begin{tabular}{cccccccccccccccccc}
\hline
&\multicolumn{13}{c}{\begin{tabular}[c]{@{}c@{}}Medical institution \\without examination capabilities$^1$\end{tabular}} & \multicolumn{4}{c}{\begin{tabular}[c]{@{}c@{}}Visit number of subject in the \\ condition of medical institution\end{tabular}} \\
&Base       & Cog       & CE       & Neur       & FB       & PE       & Blood       & Urine       & MRI       & FDG       & AV45      & Gene      & CSF      & AD                           & CN                          & Unknown                          & Total                          \\
\hline
\textbf{1}&0          & 0         & 0        & 0          & 0        & 0        & 1           & 1           & 0         & 0         & 0         & 0         & 0        & 23                           & 35                          & 952                              & 1010                           \\
\textbf{2}&0          & 0         & 0        & 0          & 0        & 0        & 1           & 1           & 0         & 1         & 0         & 1         & 1        & 0                            & 2                           & 11                               & 13                             \\
\textbf{3}&0          & 0         & 0        & 0          & 0        & 0        & 1           & 1           & 0         & 1         & 0         & 0         & 0        & 3                            & 8                           & 63                               & 74                             \\
\textbf{4}&0          & 0         & 0        & 0          & 0        & 0        & 1           & 1           & 1         & 0         & 0         & 0         & 0        & 1                            & 4                           & 23                               & 28                             \\
\textbf{5}&0          & 0         & 0        & 0          & 0        & 0        & 1           & 1           & 1         & 1         & 0         & 1         & 1        & 1                            & 0                           & 1                                & 2                              \\
\textbf{6}&0          & 0         & 0        & 0          & 0        & 0        & 1           & 1           & 0         & 1         & 1         & 1         & 1        & 0                            & 2                           & 39                               & 41                             \\
\textbf{7}&0          & 0         & 0        & 0          & 0        & 0        & 1           & 1           & 1         & 0         & 0         & 1         & 1        & 0                            & 1                           & 4                                & 5                              \\
\textbf{8}&0          & 0         & 0        & 0          & 0        & 0        & 0           & 0           & 0         & 0         & 1         & 0         & 0        & 3                            & 0                           & 19                               & 22                             \\
\textbf{9}&0          & 0         & 0        & 0          & 0        & 0        & 0           & 0           & 1         & 0         & 0         & 0         & 0        & 1                            & 0                           & 3                                & 4                              \\
\textbf{10}&0          & 0         & 0        & 0          & 0        & 0        & 1           & 1           & 1         & 1         & 1         & 1         & 1        & 1                            & 1                           & 11                               & 13                             \\
\textbf{11}&0          & 0         & 0        & 0          & 0        & 0        & 1           & 1           & 1         & 1         & 0         & 0         & 0        & 0                            & 1                           & 13                               & 14                             \\
\textbf{12}&0          & 0         & 0        & 0          & 0        & 0        & 1           & 1           & 0         & 0         & 1         & 1         & 1        & 0                            & 0                           & 26                               & 26                             \\
\textbf{13}&0          & 0         & 0        & 0          & 0        & 1        & 1           & 1           & 1         & 0         & 0         & 0         & 0        & 0                            & 0                           & 1                                & 1                              \\
\textbf{14}&0          & 0         & 0        & 0          & 0        & 0        & 0           & 0           & 0         & 1         & 1         & 0         & 0        & 0                            & 0                           & 69                               & 69                             \\
\textbf{15}&0          & 0         & 0        & 0          & 0        & 0        & 0           & 0           & 1         & 1         & 0         & 0         & 0        & 0                            & 0                           & 1                                & 1                              \\
\textbf{16}&0          & 0         & 0        & 0          & 0        & 0        & 1           & 1           & 0         & 0         & 0         & 1         & 1        & 0                            & 0                           & 19                               & 19                             \\
\textbf{17}&0          & 0         & 0        & 0          & 0        & 0        & 1           & 1           & 0         & 0         & 1         & 0         & 0        & 0                            & 1                           & 6                                & 7                              \\
\textbf{18}&0          & 0         & 1        & 0          & 0        & 0        & 1           & 1           & 0         & 0         & 0         & 1         & 1        & 0                            & 0                           & 1                                & 1                              \\
\textbf{19}&0          & 0         & 1        & 0          & 0        & 0        & 1           & 1           & 0         & 0         & 0         & 0         & 0        & 0                            & 0                           & 2                                & 2                              \\
\textbf{20}&0          & 0         & 0        & 0          & 0        & 0        & 1           & 1           & 0         & 0         & 1         & 1         & 0        & 0                            & 0                           & 5                                & 5                              \\
\textbf{21}&0          & 0         & 0        & 0          & 0        & 0        & 0           & 1           & 0         & 0         & 0         & 0         & 0        & 12                           & 45                          & 85                               & 142                            \\
\textbf{22}&0          & 0         & 0        & 0          & 0        & 0        & 1           & 1           & 0         & 1         & 1         & 0         & 1        & 0                            & 3                           & 21                               & 24                             \\
\textbf{23}&0          & 0         & 0        & 0          & 0        & 0        & 0           & 0           & 0         & 0         & 0         & 1         & 0        & 1                            & 0                           & 0                                & 1                              \\
\textbf{24}&0          & 0         & 0        & 0          & 0        & 0        & 1           & 1           & 0         & 1         & 0         & 0         & 1        & 0                            & 4                           & 10                               & 14                             \\
\textbf{25}&0          & 0         & 0        & 0          & 0        & 0        & 1           & 1           & 0         & 0         & 1         & 0         & 1        & 1                            & 0                           & 9                                & 10                             \\
\textbf{26}&0          & 0         & 0        & 0          & 0        & 0        & 0           & 1           & 0         & 1         & 1         & 0         & 1        & 0                            & 0                           & 60                               & 60                             \\
\textbf{27}&0          & 0         & 0        & 0          & 0        & 0        & 0           & 1           & 0         & 0         & 1         & 0         & 1        & 0                            & 0                           & 54                               & 54                             \\
\textbf{28}&0          & 0         & 0        & 0          & 0        & 0        & 0           & 0           & 0         & 0         & 1         & 0         & 1        & 0                            & 0                           & 1                                & 1                              \\
\textbf{29}&0          & 0         & 0        & 0          & 0        & 0        & 0           & 0           & 0         & 1         & 1         & 0         & 1        & 0                            & 0                           & 3                                & 3                              \\
\textbf{30}&0          & 0         & 0        & 0          & 0        & 0        & 0           & 1           & 0         & 0         & 0         & 0         & 1        & 0                            & 0                           & 5                                & 5                              \\
\textbf{31}&0          & 0         & 0        & 0          & 0        & 0        & 1           & 1           & 0         & 0         & 0         & 0         & 1        & 0                            & 0                           & 9                                & 9                              \\
\textbf{32}&0          & 0         & 0        & 0          & 0        & 0        & 0           & 1           & 0         & 1         & 1         & 0         & 0        & 0                            & 0                           & 1                                & 1                              \\
\textbf{33}&0          & 0         & 0        & 0          & 0        & 0        & 1           & 1           & 0         & 1         & 1         & 0         & 0        & 0                            & 0                           & 1                                & 1                              \\
\textbf{34}&0          & 0         & 0        & 0          & 0        & 0        & 1           & 1           & 0         & 1         & 1         & 1         & 0        & 0                            & 0                           & 2                                & 2                              \\
\textbf{35}&0          & 0         & 0        & 0          & 0        & 0        & 0           & 0           & 0         & 0         & 1         & 1         & 0        & 0                            & 0                           & 5                                & 5                              \\
\textbf{36}&0          & 0         & 0        & 0          & 0        & 0        & 0           & 0           & 0         & 1         & 0         & 0         & 0        & 0                            & 0                           & 1                                & 1                              \\
\textbf{37}&0          & 0         & 0        & 0          & 0        & 0        & 0           & 1           & 0         & 0         & 1         & 0         & 0        & 0                            & 0                           & 1                                & 1                              \\
\textbf{38}&0          & 0         & 0        & 0          & 0        & 0        & 1           & 0           & 0         & 1         & 0         & 0         & 0        & 0                            & 0                           & 1                                & 1                              \\
\textbf{39}&0          & 0         & 0        & 0          & 0        & 0        & 1           & 1           & 1         & 1         & 0         & 0         & 1        & 0                            & 0                           & 1                                & 1                              \\
\textbf{40}&0          & 0         & 0        & 0          & 0        & 0        & 1           & 0           & 0         & 1         & 1         & 0         & 0        & 0                            & 0                           & 1                                & 1  \\
\hline                           
\end{tabular}
\begin{tablenotes}
        \footnotesize
        \item[1] The examination is marked as 1, meaning that the medical institution cannot perform this examination for the subject. The examination is marked as 0, indicating that (1) the medical institution can perform this examination for the subject, or (2)  OpenClinicalAI does not request for performing this examination during the diagnosis of the subject though the medical institution may not be able to perform this examination for the subject. It is worth noting that the examination ability in the test set may be different from other AI systems since 0 may mean that OpenClinicalAI does not request for performing this examination during the diagnosis of the subject. However, the medical institution may not be  able to perform this examination for the subject.
  \end{tablenotes}
\end{threeparttable}
\end{table}
\clearpage

\begin{table}
\centering
\caption{\textbf{SNPs relate to AD.}\label{snp_ad}}
\begin{tabular}{cccc}
\hline
SNP\_NAME   & SNP\_NAME   & SNP\_NAME   & SNP\_NAME   \\
\hline
rs429358    & rs7412      & rs10948363  & rs7274581   \\
rs17125944  & rs4147929   & rs6656401   & rs11771145  \\
rs6733839   & rs983392    & rs10498633  & rs28834970  \\
rs9271192   & rs35349669  & rs9331896   & rs1476679   \\
rs10792832  & rs2718058   & rs190982    & rs10838725  \\
rs11218343  & rs4844610   & rs10933431  & rs9271058   \\
rs75932628  & rs9473117   & rs12539172  & rs10808026  \\
rs73223431  & rs3740688   & rs7933202   & rs3851179   \\
rs17125924  & rs12881735  & rs3752246   & rs6024870   \\
rs7920721   & rs138190086 & rs4723711   & rs4266886   \\
rs61822977  & rs6733839   & rs10202748  & rs115124923 \\
rs115675626 & rs1109581   & rs17265593  & rs2597283   \\
rs1476679   & rs78571833  & rs12679874  & rs2741342   \\
rs7831810   & rs1532277   & rs9331888   & rs7920721   \\
rs3740688   & rs7116190   & rs526904    & rs543293    \\
rs11218343  & rs6572869   & rs12590273  & rs7145100   \\
rs74615166  & rs2526378   & rs117481827 & rs7408475   \\
rs3752246   & rs7274581   &             &            \\
\hline
\end{tabular}
\end{table}

\clearpage

% \usepackage{multirow}
\begin{table}
\centering
\caption{\textbf{The normal range of indicators.}\label{index_nomal}}
\begin{threeparttable}
\begin{tabular}{cccccc}
\hline
                                                                                                        &                                                                       & \multicolumn{2}{c}{AD\_Normal} & \multicolumn{2}{c}{CN\_Normal} \\
                                                                                                        &                                                                       & Low            & High          & Low            & High          \\
 \hline
\multirow{2}{*}{\begin{tabular}[c]{@{}c@{}}Medical \\ history\end{tabular}}                             & Psychiatric                                                           & 0              & 0             & 0              & 0             \\
                                                                                                        & \begin{tabular}[c]{@{}c@{}}Neurologic \\ (other than AD)\end{tabular} & 0              & 0             & 0              & 0             \\
\multirow{2}{*}{Symptoms\tnote{1} }                                                                               & Present\_count\_21\tnote{2}                                                      & 0              & 6             & 0              & 6             \\
                                                                                                        & Present\_count\_28\tnote{3}                                                      & 0              & 8             & 0              & 8             \\
\multirow{2}{*}{\begin{tabular}[c]{@{}c@{}}Cognitive \\ Change Index\tnote{4} \end{tabular}}                      & Score\_12\tnote{5}                                                               & 32.2188        & 60            & 12             & 13.5634       \\
                                                                                                        & Score\_20\tnote{6}                                                               & 50.3438        & 100           & 20             & 22.0845       \\
\multicolumn{2}{c}{CDRSB\tnote{7} }                                                                                                                                                       & 2              & 18            & 0              & 0             \\
\multirow{3}{*}{\begin{tabular}[c]{@{}c@{}}Alzheimer's Disease \\ Assessment Scale\tnote{8} \end{tabular}}        & ADAS11\tnote{9}                                                                & 10             & 70            & 0              & 11.264        \\
                                                                                                        & ADAS13\tnote{10}                                                                & 18             & 85            & 0              & 17.67         \\
                                                                                                        & ADASQ4                                                                & 5              & 10            & 0              & 6             \\
\multicolumn{2}{c}{MMSE\tnote{11}  }                                                                                                                                                        & 0              & 27            & 25             & 30            \\
\multicolumn{2}{c}{MOCA\tnote{12}  }                                                                                                                                                        & 0              & 23            & 26             & 30            \\
\multirow{2}{*}{\begin{tabular}[c]{@{}c@{}}Preclinical Alzheimer's \\ Cognitive Composite\tnote{13}  \end{tabular}} & mPACCdigit                                                            & -30.0745       & -7.6955       & -5.1733        & 4.7304        \\
                                                                                                        & mPACCtrailsB                                                          & -29.7277       & -6.7798       & -4.8523        & 4.3338     \\
 \hline                                                                                              
\end{tabular}
\begin{tablenotes}
        \footnotesize
        \item[1] Nausea, Vomiting, Diarrhea, Constipation, Abdominal discomfort, Sweating, Dizziness, Low energy, Drowsiness, Blurred vision, Headache, Dry mouth, Shortness of breath, Coughing, Palpitations, Chest pain, Urinary discomfort (e.g., burning), Urinary frequency, Ankle swelling, Muscloskeletal pain, Rash, Insomnia, Depressed mood, Crying, Elevated mood, Wandering, Fall, Other.
        \item[2] Nausea to Rash
        \item[3] Nausea to Other
        \item[4] The CCI scale is in \url{https://adni.bitbucket.io/reference/cci.html}.
        \item[5] CCI1 to CCI12
        \item[6] CCI1 to CCI20
        \item[7] The CDR scale is in \url{https://adni.bitbucket.io/reference/cdr.html}.
        \item[8] The Alzheimer's Disease Assessment Scale-Cognitive scale is in \url{https://adni.bitbucket.io/reference/adas.html}.
        \item[9]  Q1 to Q11
        \item[10] Q1 to Q13
        \item[11] The Mini Mental State Exam scale is in \url{https://adni.bitbucket.io/reference/mmse.html}.
        \item[12] The Montreal Cognitive Assessment scale is in \url{https://adni.bitbucket.io/reference/moca.html}.
        \item[13] The calculation method of Preclinical Alzheimer's Cognitive Composite  is in \url{https://ida.loni.usc.edu/pages/access/studyData.jsp?categoryId=16&subCategoryId=43}.
  \end{tablenotes}
\end{threeparttable}
\end{table}

\clearpage

\floatname{algorithm}{Algorithm S}
\setcounter{algorithm}{0}

\begin{algorithm}
%\scriptsize
         \renewcommand{\algorithmicrequire}{\textbf{Input:}}
         \renewcommand{\algorithmicensure}{\textbf{Output:}}
         \caption{\textbf{The examination label algorithm.}\label{alg1}}
         %\label{alg1}
         \begin{algorithmic}[1]
                   \REQUIRE The label set $y_{true}$, the prediction set $y_{pred}$, diagnosis strategy set $exam\_strategy$ for a subject in a visit.
                   \ENSURE Next examination set $next\_exam$
                   \STATE Sort the $exam\_strategy$ by the number of examinations in a diagnosis strategy.
                   \FOR{$ i=0  \quad to  \quad len($exam\_strategy$) $}
                   \FOR{$ j=i+1  \quad to  \quad len($exam\_strategy$) $}
                   \IF{$exam\_strategy$[i]$\subset$$exam\_strategy$[j]}
                    \STATE  $gain=sum(y_{true}[j]\times y_{pred}[j]-y_{true}[i]\times y_{pred}[i])+sum(\sim y_{true}[i]\times y_{pred}[i]-\sim y_{true}[j]\times y_{pred}[j])$
                \IF{$gain>0$}
                \STATE  The next examination of current examination strategy $exam\_strategy$[i] is label by $exam\_strategy$[j].
                \ENDIF
                    \ENDIF
                   \ENDFOR
                   \ENDFOR
         \end{algorithmic}
\end{algorithm}

\clearpage

\begin{algorithm}
%\scriptsize
         \renewcommand{\algorithmicrequire}{\textbf{Input:}}
         \renewcommand{\algorithmicensure}{\textbf{Output:}}
         \caption{\textbf{The modified OpenMax algorithm.}\label{alg2}}
         %\label{alg1}
         \begin{algorithmic}[1]
                   \REQUIRE The abnormal pattern dataset $X$, the FitHigh function from libMR~\cite{Scheirer_2011_TPAMI}, the MiniBatchKMeans function from scikit-learn~\cite{sculley2010web}, the number of the center of known categories of subject $N$, quantiles $Q$.
                   \ENSURE The centers of known categories of subject $C$, and libMR models $Model$, the threshold of known categories of subject $Thr$.
                   \STATE $X[i]$ is the abnormal pattern dataset of $ith$ known categories of subject, in which every data $x\in X$ is belong to $ith$ known categories of subject and is correctly classified by the trained  AI model. $L$ is the number of the known categories of subject.
                   \FOR{$ i=0 \quad to  \quad (L-1)$}
                   \STATE $C[i]=MiniBatchKMeans(X[i],N[i])$
                   \ENDFOR
                   \STATE $Dist=[]$
                   \FOR{$ i=0 \quad to  \quad (L-1)$}
                   \FOR{$ x \quad in  \quad X[i]$}
                   \STATE 
                   $Dist[i]$.add(distance(x,$C[i]$,$C_{others}$) \quad // $distance=sqrt(min\_distance(x,C[i])^2+(1-min\_distance(x,C_{others}))^2)$
                   \ENDFOR
                   \ENDFOR
                   \FOR{$ i=0 \quad to  \quad (L-1)$}
                   \STATE $Model[i]$=FitHigh($Dist[i]$)
                   \STATE $Thr[i]$ is the $Q[i]$ quantile of the $Dist[i]$
                   \ENDFOR
                   \STATE Return $C$, $Model$, $Thr$
         \end{algorithmic}
\end{algorithm}

\clearpage

\begin{algorithm}
%\scriptsize
         \renewcommand{\algorithmicrequire}{\textbf{Input:}}
         \renewcommand{\algorithmicensure}{\textbf{Output:}}
         \caption{\textbf{OpenMax probability estimation.}\label{alg3}}
         %\label{alg1}
         \begin{algorithmic}[1]
                   \REQUIRE Abnormal pattern of the subject $X=\{x_1, x_2, ..., x_n\}$, raw data of subject $Z$, activation vector $V(Z)=\{v_1(Z), v_2(Z)\}$, The centers of known categories of subject $C$, and libMR models $Model$, the threshold of known categories of subject $Thr$, flag $F$, the numer of “top” classes to revise $\alpha$.
                   \ENSURE The prediction probability $\hat{P}$.
                   \STATE $L$ is the number of the known categories of subject.
                   \STATE Let $s(i)=argsort(v_j(Z))$
                   \STATE Let $Dist=[]$
                   \FOR{$ i=0\quad to  \quad (L-1)$}
                   \STATE $dist[i]=distance(X,C[i],C_{others})$
                   \ENDFOR
                    \FOR{$ i=1\quad to  \quad \alpha$}
                    %\STATE $\omega_{s(i)}(y)=1-\frac{2-i}{2}e^{-(\frac{\parallel y-\tau_{s(i)}\parallel}{\lambda_{s(i)}})^{\kappa_{s(i)}}}$
                   \STATE $\omega_{i}(Z)=1-\frac{\alpha-i}{\alpha}*models[i-1].w\_score($dist$[i-1])$
                   \ENDFOR
                   \STATE Revise activation vector $\hat{V}(Z) = V(Z)\circ\omega(Z)$
                   \STATE Define $\hat{v_0}(Z)=\sum_iv_i(Z)(1-\omega_i(Z))$
                   \STATE $\hat{P}(y=j\mid Z)=\frac{e^{\hat{v}_j(Z)}}{\sum_{i=0}^2e^{\hat{v}_i(Z)}}$
                   \IF{$F$}
                    \STATE $abnor\_score=[]$
                    \FOR {$ j=1\quad to  \quad (L-1)$}
                    \STATE $diff=dist[j-1]-thr[j-1]$
                    \IF{$diff<=0$}
                    \STATE  $abnor\_score$.append(0)
                    \ELSE
                     \STATE $tmp\_abnor\_score=diff/thr[j-1]$
                     \IF{$tmp\_abnor\_score>1$}
                     \STATE $tmp\_abnor\_score=1$
                     \ENDIF
                      \STATE $abnor\_score$.append($tmp\_abnor\_score$)
                    \ENDIF
                    \ENDFOR
                    \FOR {$ j=1\quad to  \quad (L-1)$}
                     \STATE $\hat{P}(y=j\mid Z)=\hat{P}(y=j\mid Z)*(1-abnor\_score[j-1])$
                    \ENDFOR
                     \STATE$\hat{P}(y=0\mid Z)=1-\sum_{j=1}^{L-1}\hat{P}(y=j\mid Z)$
                    \ENDIF
                   \STATE Return $\hat{P}$
         \end{algorithmic}
\end{algorithm}

\clearpage

\begin{algorithm}
%\scriptsize
         \renewcommand{\algorithmicrequire}{\textbf{Input:}}
         \renewcommand{\algorithmicensure}{\textbf{Output:}}
         \caption{\textbf{The prediction algorithm.}\label{alg_2}}
         %\label{alg1}
         \begin{algorithmic}[1]
                   \REQUIRE The base information $data_{base}$ and history recodes $data_h$ for a subject in a visit,  the trained model $model$. The threshold $\delta$, and $\gamma$.
                   \ENSURE The label of the subject.
                   \STATE $data_{input}$=$data_h$ concatenates $data_{base}$
                   \WHILE {True}
                   \STATE $result_{pred}$, $next\_examination_{pred}$=$model.predict(data_{input})$
                   \FOR{$ i=0  \quad to  \quad len( result_{pred}) $}
                   \IF{$result_{pred}[i]>=\delta[i]$}
                    \STATE Return i    \qquad  // When $i==len(result_{pred})-1$, the result is representing unknown
                   \ENDIF
                   \ENDFOR
                   \STATE $is\_concat\_new\_data=False$
                   \FOR{$ i=0  \quad to  \quad len(next\_examination_{pred}) $}
                  \IF{$next\_examination_{pred}[i]>=\gamma[i]$}
                  \IF{The $ith$ examination is able to execute by medical institution}
                  \STATE $data_{input}$=$data_{input}$ concat $data_{ith}$
                  \STATE $is\_concat\_new\_data=True$
                  %\STATE
                  \ENDIF
                  \ENDIF
                \ENDFOR
                \IF{not $is\_concat\_new\_data$}
                \STATE Select a less cost and common examination $jth$ examination which do not execute in this visit and is able to execute by medical institution.
                \IF{$jth$ examination is selected}
                \STATE $data_{input}$=$data_{input}$ concat $data_{jth}$
                \STATE $is\_concat\_new\_data=True$
                \ENDIF
                \ENDIF
                \IF{not $is\_concat\_new\_data$}
                 \STATE Return unknown
                \ENDIF
                   \ENDWHILE
                   %\STATE Return Unknown
         \end{algorithmic}
\end{algorithm}

\end{document}